\documentclass[journal]{IEEEtran}
\usepackage{amsmath}
\usepackage{algorithm}
\usepackage{algorithmic}
\usepackage{graphicx}
\usepackage{multirow}
\usepackage{tabularx}
\usepackage{xcolor}
\usepackage{array}

\usepackage{subfigure}
\ifCLASSINFOpdf
\else
\fi
\begin{document}
\title{ROSA: Robust Salient Object Detection against\\
Adversarial Attacks}

\author{Haofeng~Li,
        Guanbin~Li,
        Yizhou~Yu

        % Michael~Shell,~\IEEEmembership{Member,~IEEE,}
        % John~Doe,~\IEEEmembership{Fellow,~OSA,}
        % and~Jane~Doe,~\IEEEmembership{Life~Fellow,~IEEE}% <-this % stops a space
\thanks{This work was supported in part by the National Natural Science Foundation of China (NSFC) under Grant No. 61702565 and Grant No. U1811463, and in part by the Science
and Technology Planning Project of Guangdong Province under Grant No. 2017B010116001. (\textit{Corresponding author: Guanbin Li}.)}

\thanks{H. Li and Y. Yu are with the Department
of Computer Science, The University of Hong Kong, Hong Kong, and Y. Yu is also with Deepwise AI Lab (e-mail: lhaof@foxmail.com; yizhouy@acm.org).}% <-this % stops a space
\thanks{G. Li is with the school of Data and Computer Science, Sun Yat-sen University, Guangzhou, 510006, China (e-mail: liguanbin@mail.sysu.edu.cn).}}


\markboth{Journal of \LaTeX\ Class Files,~Vol.~14, No.~8, August~2015}%
{Shell \MakeLowercase{\textit{et al.}}: Bare Demo of IEEEtran.cls for IEEE Journals}



\maketitle

\begin{abstract}
Recently salient object detection has witnessed remarkable improvement owing to the deep convolutional neural networks which can harvest powerful features for images. In particular, state-of-the-art salient object detection methods enjoy high accuracy and efficiency from fully convolutional network (FCN) based frameworks which are trained from end to end and predict pixel-wise labels. However, such framework suffers from adversarial attacks which confuse neural networks via adding quasi-imperceptible noises to input images without changing the ground truth annotated by human subjects. To our knowledge, this paper is the first one that mounts successful adversarial attacks on salient object detection models and verifies that adversarial samples are effective on a wide range of existing methods. Furthermore, this paper proposes a novel end-to-end trainable framework to enhance the robustness for arbitrary FCN-based salient object detection models against adversarial attacks. The proposed framework adopts a novel idea that first introduces some new generic noise to destroy adversarial perturbations, and then learns to predict saliency maps for input images with the introduced noise. Specifically, our proposed method consists of a segment-wise shielding component, which preserves boundaries and destroys delicate adversarial noise patterns and a context-aware restoration component, which refines saliency maps through global contrast modeling. Experimental results suggest that our proposed framework improves the performance significantly for state-of-the-art models on a series of datasets.
\end{abstract}

\begin{IEEEkeywords}
Deep Neural Network, Adversarial Attack, Salient Object Detection.
\end{IEEEkeywords}

% \fi
\IEEEpeerreviewmaketitle

\section{Introduction}
\IEEEPARstart{S}{alient} object detection aims at locating and segmenting objects, which are most visually distinctive to human subjects, in an image or a video frame. Designing a salient object detection model for simulating this process not only improves our understanding of the inner mechanism of human vision and psychology, but also benefits many applications in the field of computer vision and graphics. For example, salient object detection has been widely studied and applied to robotics~\cite{yu2010object}, context-aware image editing~\cite{goferman2012context}, object segmentation~\cite{wei2017stc}~\cite{Wang2014VOC} and person re-identification~\cite{bi2014person}. Since salient object detection algorithms are usually adopted during the initialization or pre-processing stage of a system, efficiency and robustness are of considerable importance. Imagine if the performance of the pre-processing stage is seriously affected by corrupt input images, succeeding stages might produce unpromising results, which could be a catastrophe to the entire system.
\begin{figure}
\centering
\includegraphics[scale=1.05]{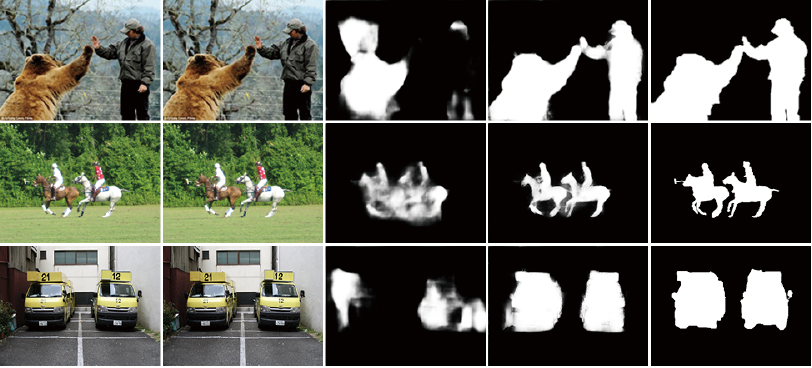}
\leftline{\footnotesize \hspace{0.35cm} Original \hspace{0.45cm} Adversarial \hspace{0.2cm} (a) DSS~\cite{HouPami18Dss} \hspace{0.07cm} (b) DSS+ours \hspace{0.3cm} Ground}
\leftline{\footnotesize \hspace{0.4cm} Images \hspace{0.7cm} Samples \hspace{4.25cm} Truth}
\caption{Effectiveness of our proposed method. The leftmost column shows original images while the second column from the left displays the corresponding adversarial samples. The $L_\infty$ norm of the adversarial perturbations is set as 25 pixel values. Column (a) is the saliency maps of adversarial samples, and is predicted by DSS~\cite{HouPami18Dss}. Column (b) is the saliency maps of adversarial samples, and is predicted by our proposed method with DSS as the backbone network stream. The rightmost column is the ground truth saliency maps. As can be seen in the above, the adversarial samples are almost visually the same as their original images. Besides, DSS incorporated with our proposed method yields saliency maps of higher quality, in comparison to the original DSS.}
\label{fig1}
\end{figure}

For the last several years, significant successes have been achieved in the computer vision community, as training deep convolutional neural networks (CNN) on large-scale datasets becomes feasible. A deep CNN is composed of stacked convolution filters with learnable parameters. Since those filters harvest information naturally from local neighborhoods in the input image and their parameters are adaptively determined by a training set, deep CNNs demonstrate a high fitting capacity superior to traditional methods using handcrafted features. Nowadays deep learning has been widely employed in image classification, semantic segmentation, object localization as well as salient object detection.

Deep learning based salient object detection models can be roughly divided into two groups. One group adopts segment-wise labeling while the other group predicts pixel-level results. Segment-wise labeling methods first divide an image into regions. Pixels in the same region most probably share similar saliency values. CNN features for each region are then extracted to evaluate its saliency. In contrast, pixel-wise methods usually embrace fully convolutional network architectures, which take a whole image as input and yield a dense saliency map directly. Such methods not only demonstrate higher efficiency but also achieve state-of-the-art accuracy in virtue of their end-to-end trainable property.

However, those FCN driven approaches have weaknesses that might degrade their performance in practice. First, being end-to-end trainable allows gradients propagated easily from supervision target to the input image, which puts the salient object detection models at the risk of adversarial attacks. Adversarial attacks generate adversarial samples that do not change the ground truth assigned by human subjects but increase the prediction error of neural models by making visually imperceptible changes to the image, as shown in Figure 1. Second, dense labeling models do not explicitly model contrast among different image parts but implicitly estimate saliency in a single FCN. Once input images are polluted by adversarial noise, low-level features and high-level features that cannot correct themselves will be affected as well. Third, current largest training datasets of salient object detection contain only several thousands of images, in comparison to some image classification benchmark with millions of samples~\cite{deng2009imagenet}. At the same time, the salient object categories included are very limited. Thus to some extent existing models are fitting bias within the data, for example, detecting objects frequently appearing in training set rather than locating the most distinctive ones. Those approaches might rely on capturing too much high-level semantics and could be sensitive to low-level perturbations, such as adversarial noises.

Segment-wise labeling approaches enjoy higher robustness as they model contrast explicitly and determine saliency score depending on multiple regions, such as the considered segment and its context. For different segments, gradients calculated by the same target might conflict when propagating on the input image, since different regions could share the same local or global context. Nevertheless, it is inefficient to adopt sparse labeling methods in practice, due to evaluating hundreds of segments.

To enhance robustness and maintain efficiency for existing dense labeling methods, this paper proposes a novel framework ROSA (named after RObust SAliency) that can take any FCN as a backbone. We first observe that adversarial noise itself is fragile as it is computed accurately by backward propagation. Adversarial noise forms some subtle curve-like pattern that may play an important role. Destructing such patterns could reduce the attack effects. And then we notice that convolutional neural networks are less sensitive to some generic noise than adversarial noises, since adversarial samples are aimed at neural models. We also consider a priori that nearby pixels with similar low-level features have similar saliency values. Thus we come up with a novel framework that first destroys adversarial perturbations by introducing some new generic noise, and then learns to adaptively predict saliency maps against the new introduced noise. To destroy adversarial noises, we develop a segment-wise shielding component placed before the backbone network. Segment-wise shielding component divides an image into small parts according to low-level similarity and shuffles pixels in each part randomly. It introduces another generic noise to destroy the structural pattern in adversarial samples and therefore alleviates the attack effect. To refine results affected by the newly introduced noise, we conceive another component known as context-ware restoration placed after the backbone network. The restoration component adjusts saliency score at some position according to similarities among raw pixel values of the position and its context. The overall system with a backbone network is fine-tuned end-to-end in the training stage.

Our proposed framework demonstrates several strengths in the following. First, the ROSA framework is not so susceptible to adversarial attacks. Since the shielding component has no learnable parameters, it does not support backward propagating gradients onto the input image to generate adversarial samples. Even when adversarial samples are found, their adversarial noise can still be destroyed by the shielding component during the testing stage. Second, the shielding component shuffles pixels in the same segment and thus does boundary less harm. Moreover, ROSA adopts a FCN based model as its backbone and the restoration component is implemented by convolutional operator that supports parallel computing. Both designs help maintain acceptable efficiency for the entire system.

In short, our contributions have three folds.
\begin{itemize}
\item We for the first time launch adversarial attacks on state-of-the-art salient object detection models successfully.
\item We propose a novel salient object detection framework that first introduces some new noise to resist adversarial perturbations, and then adaptively predicts saliency maps for inputs with the new introduced noises. The proposed framework is instantiated by an arbitrary FCN backbone and two strongly coupled and complementary components.
\item Experimental results verify that the implemented adversarial attacks are effective for a wide range of existing salient object detection models. Moreover, extensive experiments demonstrate that the our proposed framework is resistant to adversarial samples, and more robust than existing defense baselines.
\end{itemize}

\section{Related Work}
In this section we brief several groups of previous work related to our proposed approach, salient object detection, adversarial attacks and defenses against adversarial attacks.

%\IEEEpubidadjcol

\subsection{Salient Object Detection}
Algorithms for detecting salient objects can be separated into two categories. One category is the conventional methods that do not use neural networks but resort to prior knowledge and handcrafted features~\cite{cheng2013efficient,yang2013saliency,jiang2013salient,cheng2015global,huang2017300,goferman2012context,tu2016real,zhang2015minimum,zhu2014saliency,zhang2017ranking,zhang2018saliency}. Ranking saliency~\cite{zhang2017ranking} is a saliency detection algorithm based on graph-based manifold ranking, which ranks the relevances of images elements to foreground or background seeds.
Another category driven by deep convolutional neural networks can be categorized as two groups, sparse labeling and dense labeling. Sparse labeling methods~\cite{li2015visual,kim2016shape} appeared in early years. Li and Yu~\cite{li2015visual,li-mdf} trained a binary classifier to estimate visual saliency for each superpixel with multi-scale learned CNN features. Wang et al.~\cite{wang2015deep} developed a local DNN estimating coarse saliency for object proposals and a global DNN evaluating weights to combine different proposals. Zhao et al.~\cite{zhao2015saliency} employed deep CNN to predict visual saliency for single superpixel with local context and global context. Qin et al.~\cite{qin2018hierarchical} introduce Single-layer Cellular Automata (SCA) which can exploit the intrinsic relevance of similar image regions to detect salient objects, based on extracted deep features.
Since these methods take a region as a unit of computation and contain two separate steps of feature extraction and salient value inference, they are generally inefficient and require a large amount of space for feature storage. Inspired by the successful application of fully convolutional networks in pixel-level semantic segmentation, recently dense labeling approaches have established the new state-of-the-art in salient object detection~\cite{li-dcl,li-instance,liweaksal,luo2017non,zhang2017amulet,zhang2017learning,wang2018detect,chen2018reverse,Chen2018Emb,wang2018salient}. Li and Yu~\cite{LiYu16} modeled visual saliency by combining a fully convolutional stream with a segment-wise spatial pooling stream. Wang et al.~\cite{Wang2016Saliency} employed fully convolutional networks to refine coarse saliency maps based on prior knowledge in a recurrent way. Hou et al.~\cite{HouPami18Dss} adapted Holistically-Nested Edge Detector (HED)~\cite{xie2015holistically} architecture by introducing short connections to the skip-layer structures.

\subsection{Adversarial Attack}
 Existing adversarial attacks consist of several groups, one-step gradient-based methods~\cite{Goodfellow2015Explaining}, iterative methods~\cite{Dong_2018_CVPR,moosavi2016deepfool,NIPS2017_7273,kurakin2017adversarial,liu2017delving,Moosavi-Dezfooli_2017_CVPR,madry2018towards}, optimization-based methods~\cite{nguyen2015deep,xiao2018spatially} and generative networks~\cite{Poursaeed_2018_CVPR,zhao2018generating} based methods. The fast gradient sign method (FGSM)~\cite{Goodfellow2015Explaining} computes one-step gradient to maximize the loss $L(\cdot)$ between model output and the ground truth, within some $L_\infty$ norm bound $\epsilon$. FGSM generates adversarial sample as in Equation~\ref{eq1}:
\begin{equation} \label{eq1}
x^* = x + \epsilon \cdot sign(\bigtriangledown_xL(f(x;\theta),y))
\end{equation}
where $x^*$, $x$, $y$ are the adversarial sample, original image, and ground truth respectively. $f(\cdot; \theta)$ denotes some neural model with parameters $\theta$. Iterative approaches~\cite{kurakin2017adversarial,madry2018towards} conduct FGSM multiple times with a small step length $\alpha$ as Equation~\ref{eq2} shows:
\begin{equation} \label{eq2}
x^*_{t+1} = clip(x^*_t + \alpha \cdot sign(\bigtriangledown_xL(f(x;\theta),y)),\epsilon)
\end{equation}
where $x^*_t$ denotes an adversarial sample obtained at $t$-th time step. $x^*_0$ is initialized as $x$. $clip(x,\epsilon)$ keeps each element $x_i$ of $x$ within the range of $[x_i-\epsilon, x_i+\epsilon]$. Szegedy et al.~\cite{szegedy2014intriguing} solved a box-constrained optimization with L-BFGS to find an adversarial sample. Y. Dong et al.~\cite{Dong_2018_CVPR} proposed an iterative algorithm that integrates a momentum term into the iterative process to boost adversarial attacks. Moustapha Cisse et al.~\cite{NIPS2017_7273} proposed an approach called `Houdini' to attack structured prediction problems (including human pose estimation and speech recognition) whose final performance measure is a combinatorial non-decomposable quantity. Dai, Hanjun et al.~\cite{pmlr-v80-dai18b} proposed to fool a family of graph neural networks by modifying the combinatorial structure of data. They developed a reinforcement learning based methods, variants of genetic algorithms and gradient based methods to attack graph neural networks. Adversarial attack on salient object detection remains a gap before this paper.

\subsection{Defense against Adversarial Attacks}
Some defense methods are proposed to protect attacked target neural networks form potential adversarial samples~\cite{papernot2016distillation,metzen2017on,lu2017safetynet,guo2018countering,xie2018mitigating,song2018pixeldefend,Liao_2018_CVPR,Akhtar_2018_CVPR}.
 J. H. Metzen et al.~\cite{metzen2017on} augmented the attacked target network by small subnetworks, which take output feature maps at some layers as inputs, and predict a probability of the input containing adversarial noise. SafetyNet~\cite{lu2017safetynet} equips a convolutional neural network classifier with an RBF-SVM to detect adversarial samples with discrete codes calculated from the final RELU outputs. Images transformations including bit quantization, vectorization~\cite{Wang2017VVV}, JPEG compression and total variance minimization may remove or destroy adversarial perturbations~\cite{guo2018countering}, before feeding an input image into the target network. C. Xie et al.~\cite{xie2018mitigating} proposed a simple method that randomly resizes an input image and pads it with zeros, to destroy the effect of adversarial attacks. F. Liao et al.~\cite{Liao_2018_CVPR} developed a neural network based denoiser that is trained with a loss function based on some high-level features of the attacked target classifier. Many existing defense baselines struggle to remove potential adversarial noises from input images, which is different from our proposed idea that adaptively predicts saliency maps for inputs with some new introduced generic noise to resist adversarial attacks.

\begin{figure*}
\flushleft
\includegraphics[scale=1.04]{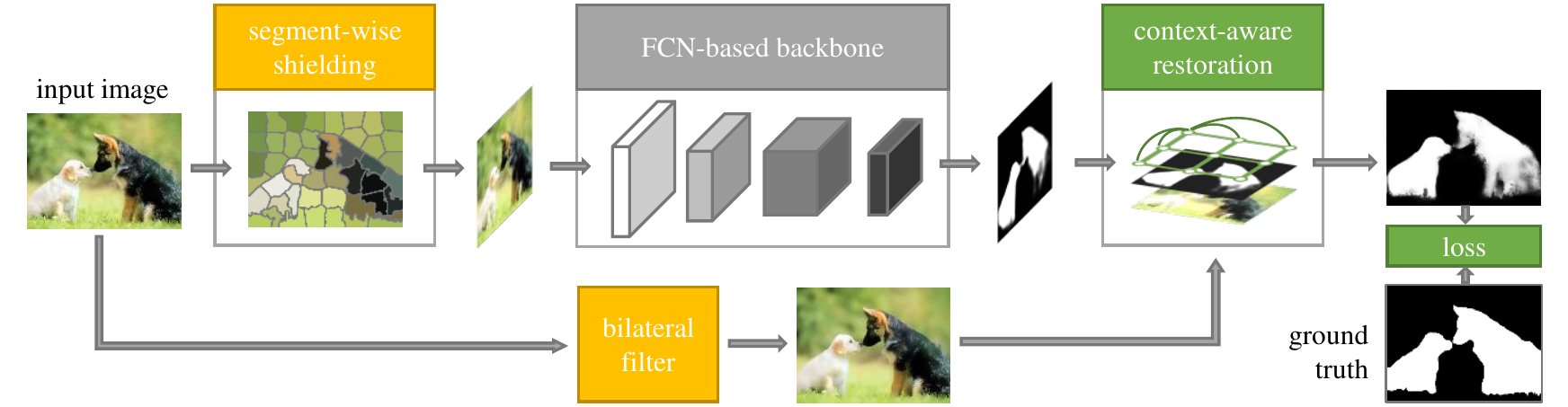}
\caption{ROSA, Robust Salient Object Detection Framework. To defend against adversarial samples, the proposed method exploits a novel idea that first introduces some novel generic noise to destroy adversarial perturbations, and then learns to predict saliency maps for images with the introduced generic noise. Such idea is different from those resorting to image smoothing or transformations before feeding input images into target networks. As shown in the above, the segment-wise shielding component introduces some generic noise to perturb the adversarial noises. Then a fully convolutional network (FCN) based backbone takes the noisy image as input, and yields a coarse saliency map. Afterwards, a context-aware restoration component utilizes a graph model to refine the coarse saliency map, with a smoothed image providing pairwise pixel-level similarity. The smoothed image is obtained by applying bilateral filter on the input image. Lastly, a pixel-wise binary cross-entropy loss function is calculated between the refined saliency map and the ground truth. The FCN-based backbone network and the context-aware restoration component are end-to-end trained to adapt to the input images with the introduced noise. The proposed method can theoretically incorporate arbitrary FCN-based backbone network.}
\label{fig2}
\end{figure*}
\section{Method}
This section first describes how we launch an adversarial attack on state-of-the-art visual saliency models, and then detail how our proposed robust salient object detection framework works.

\subsection{Adversarial Samples for Salient Object Detection}
Adversarial attack aims at synthesizing some perturbed input that fools neural models without changing its ground truth label. In this section, we introduce the pipeline to yield an adversarial sample for a given salient object detection model $f$, and it can be directly used to attack other detection models as most of the existing visual saliency models have similar FCN-based network architectures and are usually initialized by the same pre-trained image classification model~\cite{he2016deep,simonyan2014very}. Adversarial samples can be divided into two categories. \textit{Targeted} adversarial samples make attacked models produce specific results as predicted saliency maps while \textit{non-targeted} ones maximize mean absolute error (MAE) and/or minimize $F_\beta$ measure. In this paper, only non-targeted attacks are concerned and targeted samples may be investigated in the future.

Inspired by~\cite{xie2017adversarial}, we implement an iterative gradient-based pipeline to synthesize adversarial samples. To generate those samples, it requires a neural network pre-trained on salient object detection, some natural images and their corresponding saliency maps densely labeled at pixel level. Let $f(\cdot,\theta)$ be the pre-trained model with parameters $\theta$. $x$, $x^*$ and $y$ denote a natural image, its corresponding adversarial sample and ground truth, respectively. Before synthesizing the adversarial sample, $x$ is subtracted by mean pixel values. After the generation, $x^*$ is enlarged to the range of $[0,255]$ and rounded to RGB image. Each element $y_i$ of $y$ belongs to $\{0,1\}$, with 0 denoting non-salient and 1 denoting salient. To ensure the adversarial perturbation unnoticeable, parameter $\epsilon$ is set as upper bound of $L_\infty$ norm such that $||x-x^*||\leq \epsilon$. The maximum number of iterations $T$ limits the overall running time cost. Once $T$ iterations are finished or the $L_\infty$ norm bound is reached, the generation stops and returns adversarial sample obtained at current time step.

In each iteration $t$, supposing that adversarial sample $x^*_t$ from previous time step or initialization is prepared, we update the adversarial sample as in Equation~\ref{eq3}.
\begin{equation} \label{eq3}
x^*_0 = x, x^*_{t+1} = x^*_t + p_t
\end{equation}
where $p_t$ denotes adversarial perturbation computed at $t$-th step. We formulate the goal making the predictions of all pixels in $x$ go wrong as $\forall i, argmax_c\{f_{i,c}(x^*_t+p_t;\theta)\}\neq y_i$. Here, $i$ denotes one of all $n$ pixels in $x$ and $c$ denotes two categories: salient and non-salient. To determine $p_t$, gradient descent algorithm is applied as in Equation~\ref{eq4}
\begin{equation} \label{eq4}
p'_t = \sum\limits_{i\in S_t} [\bigtriangledown_{x^*_t}f_{i,1-y_i}(x^*_t;\theta) - \bigtriangledown_{x^*_t}f_{i,y_i}(x^*_t;\theta)]
\end{equation}
where $S_t$ denotes the set of pixels that $f$ still can classify correctly. Then $p_t$ is obtained by normalization as $\alpha \cdot p'_t / ||p'_t||_\infty$ where $\alpha$ is a fixed step length. The pseudocode of the entire generation pipeline is shown in Algorithm~\ref{algo1}.
\begin{algorithm}
\caption{Adversarial Sample Generation}
\label{algo1}
%\small
\begin{algorithmic}
\REQUIRE natural image $x$;
\STATE \hspace{\algorithmicindent} \hspace{\algorithmicindent} corresponding saliency annotation $y$;
\STATE \hspace{\algorithmicindent} \hspace{\algorithmicindent} pre-trained visual saliency model $f(\cdot;\theta)$;
\STATE \hspace{\algorithmicindent} \hspace{\algorithmicindent} pixels set $S = \{1,2,...,n\}$ of $x$;
\STATE \hspace{\algorithmicindent} \hspace{\algorithmicindent} maximum number of iterations $T$;
\STATE \hspace{\algorithmicindent} \hspace{\algorithmicindent} step length $\alpha$; upper bound $\epsilon$ of $L_\infty$ norm;

\STATE $x^*_0 \leftarrow x, p \leftarrow 0, t \leftarrow 0, e \leftarrow 0, S_0 = S$;
\WHILE{$t < T$ and $e \leq \epsilon$ and $|S_t| > 0$}
\STATE calculate $p'_t$ by Equation~\ref{eq4};
\STATE $p_t \leftarrow \alpha \cdot p'_t / ||p'_t||_\infty$;
\STATE $p \leftarrow p + p_t$;
\STATE calculate $x^*_{t+1}$ by Equation~\ref{eq3};
\STATE $e \leftarrow ||x^*_{t+1} - x||_\infty$;
\STATE $t \leftarrow t + 1$;
\STATE $S_t \leftarrow \{i | argmax_c\{f_{i,c}(x^*_t;\theta)\} = y_i\} $;
\ENDWHILE
\STATE $x^* \leftarrow x^*_t + \overline{x}$;
\STATE $x^* \leftarrow round(x^*)$;
\RETURN $x^*$;
\end{algorithmic}
\end{algorithm}

\subsection{Robust Salient Object Detection Framework}
In this section, we propose a novel salient object detection framework ROSA that demonstrates high robustness against adversarial attacks. As shown in Figure 2, the ROSA framework consists of a segment-wise shielding component, a FCN-based backbone network and a context-aware restoration component. Virtually the backbone can be chosen as an arbitrary FCN-based visual saliency model that takes a whole image as input and yields a densely labeled saliency map. The FCN backbone enjoys high efficiency and accuracy but displays sensitivity on adversarial samples. The shielding component and the restoration component play an important role in improving the robustness of the proposed framework.

A segment-wise shielding component destroys potential adversarial noise patterns in an input image before sending it to the backbone, by introducing some ``shuffling'' noise that is easier to resist. The observation behind is that adversarial noises are some delicate perturbations deliberately synthesized for convolutional neural networks, while CNN is not sensitive to and can adapt to some other noise. To alleviate harms caused by the new noise, the shielding component first divides the input image into non-overlapping regions, namely superpixels. We follow the region decomposition method developed by~\cite{Achanta2009Frequency}. Specifically, k cluster centers in the joint space of color and pixel position are initialized by sampling pixels at regular grid steps. Then we assign each pixel to the cluster center with minimum distance, and update each cluster center as the mean vector of pixels belonging to the cluster, in an iterative way. The iteration ends when L2 norm error between new location and previous location of each cluster center converges.

After region decomposition, we permute all pixels within the same superpixel randomly. Such shuffling operation strongly destroy the adversarial perturbation while it limits the introduced noise within each single superpixel. Thus object boundaries that those superpixels are adhere to are not spoiled and the noisy saliency map output by the backbone network has a chance to be restored. Some may suggest an option that smooths each superpixel by averaging pixels inside. Recall what we argue in the introduction, existing FCN models overfit too much high-level semantics in visual saliency data. The random permutation makes capturing high-level semantics more difficult and enforces neural networks to harvest low-level contrast among regions. It also plays a role in augmenting dataset and reducing the overfitting issue.

A context-aware restoration component exploits low-level similarity between each pixel and its context to refine the saliency scores provided by the backbone network. As adversarial perturbations aim at parameterized convolution filters, the restoration component adopts a complete graph model instead of CNN architecture. We measure similarity among pixels in low-level color space and spatial position, since previous high-level convolutional features have been
polluted. The restoration component adjusts saliency maps by minimizing some energy function as Equation~\ref{eq5}:
\begin{equation} \label{eq5}
E(y^*) = \sum\limits_{i}E_u(y^*_i,y_i) + \sum\limits_{i<j}E_p(y^*_i,y^*_j)
\end{equation}
where $y$ denotes the coarse saliency map and $y^*$ denotes the resulted saliency map. The first unary energy term measures the cost (inverse likelihood) of assigning $i$ with $y^*_i$. The second term pairwise energy measures the cost of assigning $i$ and $j$ with $y^*_i$ and $y^*_j$ at the same time. It encourages similar nearby pixels to be labeled the same. The pairwise energy is defined as Equation~\ref{eq6} where $p$ denotes pixel position and $x'_i$ denotes pixel color. $x'$ is a smoothed image output by the bilateral filter that takes the adversarial sample $x^*$ as input, as shown in Figure~\ref{fig2}. $\omega_1$ and $\omega_2$ are tuned by training. $\theta_\alpha$, $\theta_\beta$ and $\theta_\gamma$ are chosen as 160, 3 and 3 respectively. $\mu$ is a learnable label compatibility function that penalizes assigning $i$ and $j$ with different labels.
\begin{equation} \label{eq6}
\begin{split}
E_p(y^*_i,y^*_j) = &\mu(y^*_i,y^*_j) \{\omega_1exp(-\frac{|p_i-p_j|^2}{2\theta^2_\alpha}-\frac{|x'_i-x'_j|^2}{2\theta^2_\beta})\\
&+ \omega_2exp(-\frac{|p_i-p_j|^2}{2\theta^2_\gamma}) \}
\end{split}
\end{equation}

We realize the component following some previous work~\cite{krahenbuhl2011efficient,zheng2015conditional} that solves Equation~\ref{eq5} as densely connected conditional random field with a recurrent neural network. The neural network is implemented with and enjoys efficiency from $1\times1$ convolutional layers. Since the restoration component makes use of global context to refine results, it is more difficult to change the prediction by adversarial noises of some limited perturbation strength. In order to influence the prediction results of pixels at certain specific locations, intricate changes involving a larger range of feature vectors may be required, which in turn results in larger pixel value perturbations.

\subsection{Training Scheme}
The following explains how we train the entire framework in an end-to-end scheme. In the beginning, the FCN-based backbone of ROSA framework is initialized as some pre-trained visual saliency model while the parameters of context-aware restoration component are set up according to~\cite{zheng2015conditional}.
Then the parameters of the backbone network and the restoration component are fine-tuned together. As the segment-wise shielding component contains no learnable parameters, gradients are not passed backward through that component. To maintain generalization ability against different kinds of adversarial perturbation, our training set does not include adversarial samples but only natural images. These training samples are fed into the segment-wise shielding component. As shown in Figure~\ref{fig2}, a pixel-wise cross-entropy loss function is computed between the ground truth saliency map and the output of the context-aware restoration component. SGD algorithm is used to train the proposed method. The learning rate of the context-aware restoration component is set as $10^{-10}$ while that of other parts is selected as $10^{-13}$. The momentum and weight decay are set as 0.9 and 0.0005 respectively. For each backbone FCN in this paper, fine-tuning with our proposed framework takes no more than 5 epochs. We adopt early-stopping strategy and terminate the training if the performance on validation set is not improved after 2 consecutive epochs. If the proposed method adopts DSS as its backbone, a forward pass on an image costs about 0.8 seconds.

\section{Experiment}
In this section, we conduct three groups of experiments. First, we launch adversarial attacks on existing visual saliency models and investigate how they are affected. Then, we integrate our proposed framework with current models to present how the proposed framework enhances robustness for those models. Lastly, we verify the effectiveness of each component in the ROSA framework.

\subsection{Dataset}
In this paper, we conduct experiments on MSRA-B dataset~\cite{Liu2007Learning}, HKU-IS dataset~\cite{li2015visual}, DUT-OMRON dataset~\cite{yang2013saliency} and ECSSD dataset~\cite{shi2016hierarchical}. MSRA-B dataset contains a train set of 2500 images, a validation set of 500 images and a test set of 2000 images. HKU-IS dataset includes 2500 images, 500 images and 1447 images in train set, validation set and test set respectively. We follow the released data split in MSRA-B and HKU-IS dataset. For DUT-OMRON dataset, we randomly separate all the 5168 images into a train set of 2500 images, a validation set of 500 images and a test set of 2168 images. For ECSSD dataset, all 1000 images are taken as testing samples.
\begin{table}[!t]
\centering
\caption{$F_\beta$ measure and MAE on Natural and Adversarial Examples}
\label{table:attack_effect}
\small
\begin{tabular}{ c|c c|c c }
\hline
\multirow{2}{*}{Model} &
	\multicolumn{2}{c|}{$F_\beta$ measure} &
	\multicolumn{2}{c}{MAE} \\
\cline{2-5}
       & Original  & Adversarial & Original & Adversarial \\
\hline \hline
 DSS   & 87.13\%   & 51.08\% & 0.0491 & 0.2251 \\
 DCL   & 86.82\%   & 56.84\% & 0.0583 & 0.1921 \\
 RFCN  & 87.68\%   & 75.39\% & 0.0544 & 0.0976 \\
 Amulet& 84.94\%   & 75.63\% & 0.0511 & 0.0906 \\
 UCF   & 81.64\%   & 75.03\% & 0.0734 & 0.1107 \\
 LEGS  & 76.39\%   & 75.43\% & 0.1192 & 0.1231 \\
 MC    & 75.31\%   & 74.65\% & 0.0982 & 0.0999 \\
 MDF   & 82.05\%   & 80.99\% & 0.0946 & 0.0999 \\
\hline
\end{tabular}
\end{table}

\subsection{Evaluation}
We select Mean Absolute Error (MAE), precision, recall, $F_\beta$ measure and PR curves as evaluation metrics. MAE measures pixel-level difference between the saliency map $S$ and ground truth $G$ as Equation~\ref{eq9}:
\begin{equation} \label{eq9}
MAE = \frac{1}{W\times H}\sum\limits_{i=1}^W \sum\limits_{j=1}^H |S_{i,j}-G_{i,j}|
\end{equation}
where $W$ and $H$ denote the width and height of the saliency map respectively. To compute $F_\beta$ measure, we binarize each saliency map with an image-dependent threshold proposed by~\cite{Achanta2009Frequency}. The threshold $T$ is calculated as Equation~\ref{eq7}:
\begin{equation} \label{eq7}
T = \frac{2}{W\times H}\sum\limits_{i=1}^W \sum\limits_{j=1}^H S_{i,j}
\end{equation}
where $W$ and $H$ denote width and height of the saliency map $S$. Pixels with saliency value larger than $T$ form the predicted salient region. Precision is the ratio of ground truth salient pixels in the predicted salient area while recall is the ratio of predicted salient pixels in the ground truth salient area. $F_\beta$ measure is defined as Equation~\ref{eq8}~\cite{Achanta2009Frequency}:
\begin{equation} \label{eq8}
F_\beta = \frac{(1+\beta^2)\times Precision \times Recall}{\beta \times Precision + Recall}
\end{equation}
where $\beta^2$ is set as 0.3 to emphasize the precision. To draw PR curves, a list of equally spaced thresholds are sampled. For each threshold value, each predicted saliency map in the benchmark is quantized into a binary mask. Precision and recall are calculated with each binary mask and its ground truth annotation. The precision and the recall corresponding to each threshold are computed by respectively taking average of precision and recall for all binary masks. Then we obtain a list of (\textit{precision}, \textit{recall}) pairs and plot it as a PR curve.
\begin{figure}[!t]
\centering
\includegraphics[scale=0.32]{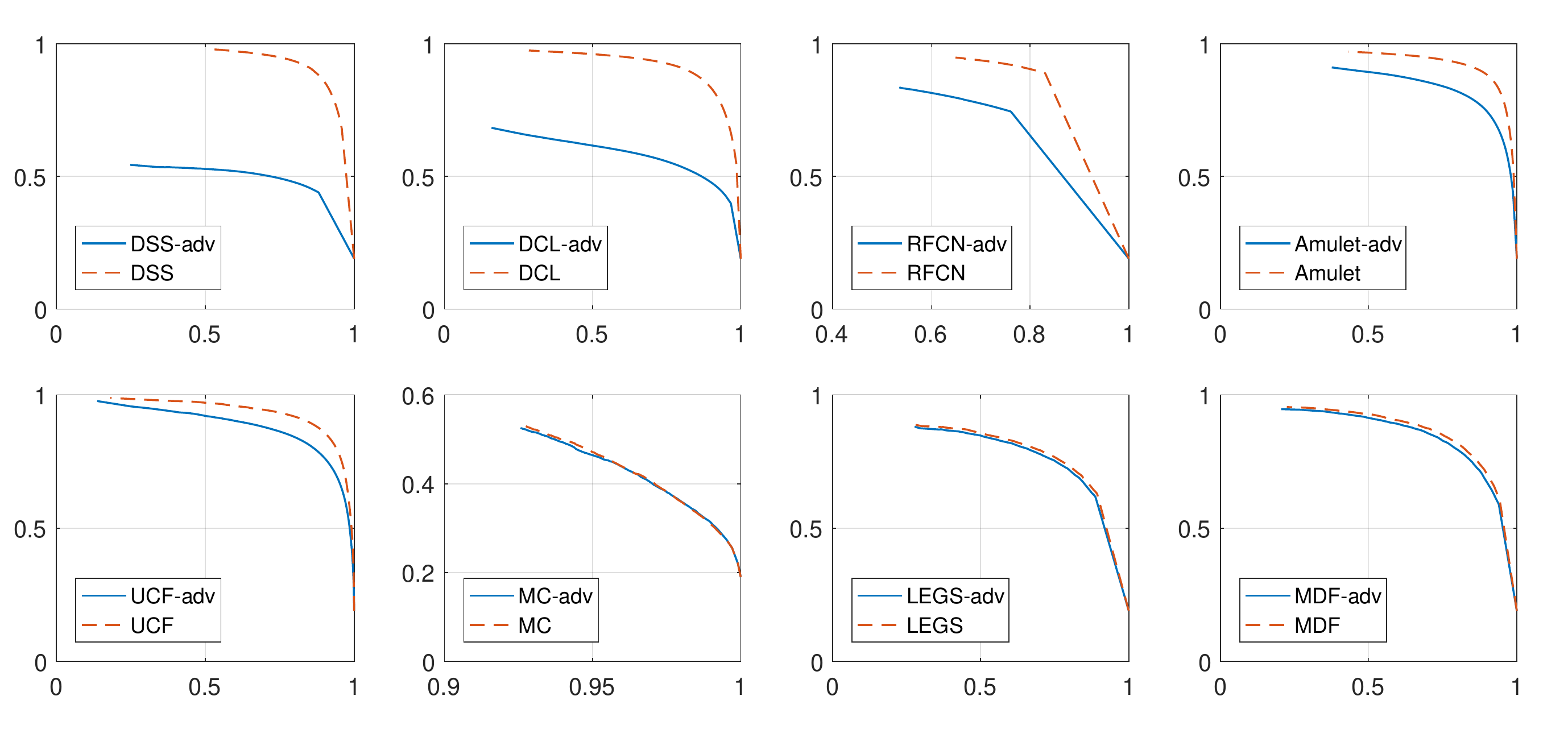}
\caption{Effectiveness of Adversarial Attack on PR curves. As shown in the above figures, the PR curve of DSS on adversarial samples drops the most seriously. The PR curves of DCL, RFCN, Amulet and UCF also degrade to some extent, which suggests that adversarial samples yielded by some FCN network are transferable to attack other FCN variants. For MC, LEGS and MDF, their PR curves tested on natural images and adversarial samples are relatively close to each other, which indicates that the sparse labeling based methods are insensitive to adversarial noises.}
\label{fig:attack_PR_curve}
\end{figure}

\subsection{Effectiveness of Adversarial Attack}
We demonstrate the performance of eight state-of-the-art visual saliency models: DSS~\cite{HouPami18Dss}, DCL~\cite{LiYu16}, RFCN~\cite{Wang2016Saliency}, Amulet~\cite{zhang2017amulet}, UCF~\cite{zhang2017learning}, MC~\cite{zhao2015saliency}, LEGS~\cite{wang2015deep} and MDF~\cite{li2015visual} on natural images and adversarial samples which are synthesized with a pre-trained DSS model. For efficiency, the above neural models are trained on the train set of MSRA-B and tested on the test set of HKU-IS. The upper bound of $L_\infty$ norm $\epsilon$ is chosen as 20. Qualitative results can be found in Figure~\ref{attackSample} where each sample consists of two rows. The upper are natural image and its predicted saliency maps while the lower are adversarial samples and their corresponding results. The second column from the left are the ground truth saliency maps denoted as GT. DSS, DCL, RFCN, Amulet and UCF. Predicted saliency maps on adversarial samples change significantly, compared with that on natural images. For MC, LEGS and MDF, predictions on adversarial samples and that on original images are visually approximate.
\begin{figure}[!t]
\centering
\includegraphics[scale=0.32]{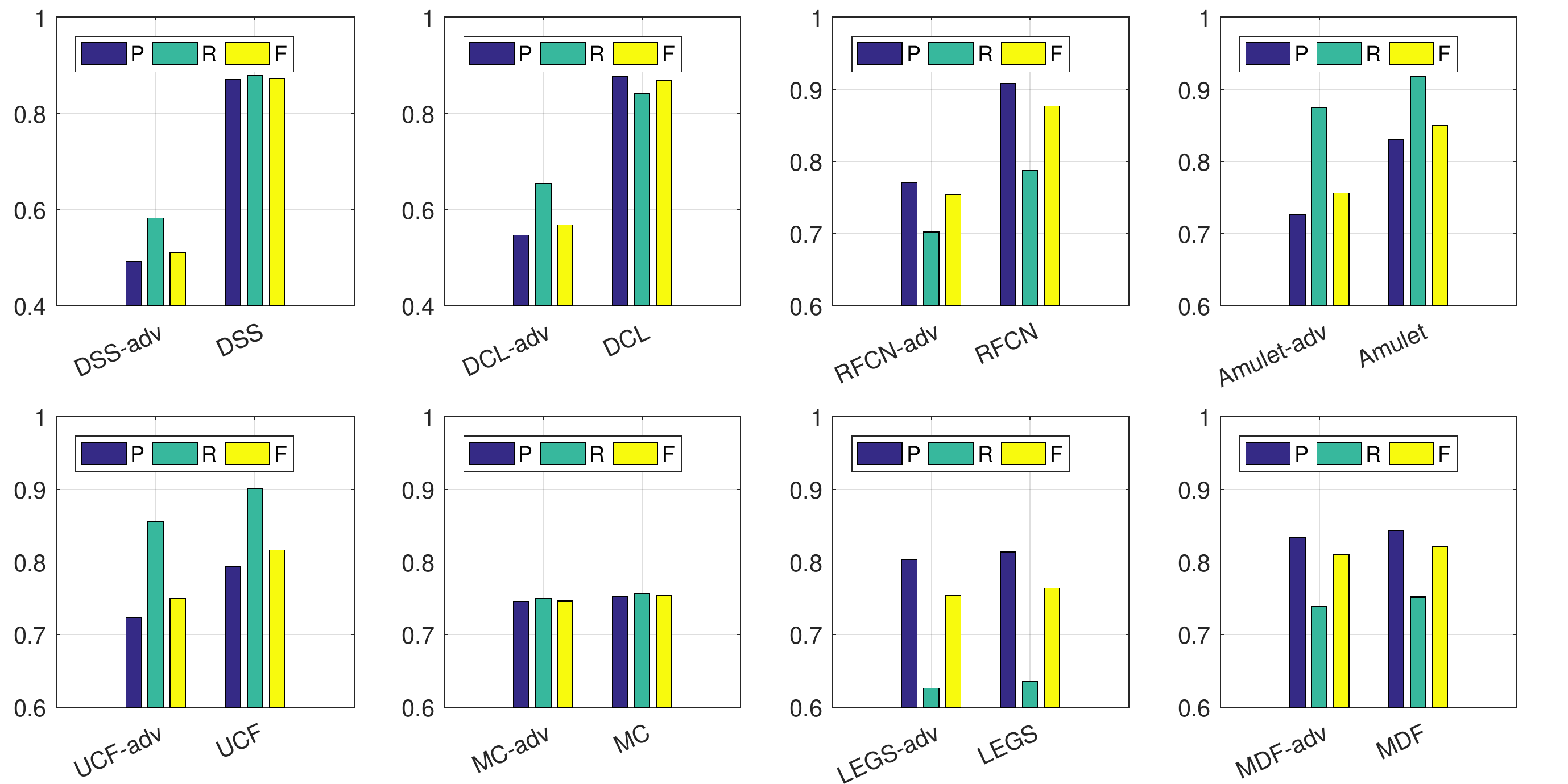}
\caption{Effectiveness of Adversarial Attack on Precision, Recall and $F_\beta$ measure. $N$-adv denote the result of neural network $N$ tested on adversarial samples. As shown in the above bar diagrams, the precision, recall and $F_\beta$ of DSS-adv and DCL-adv decline by a wide margin, in comparison to DSS and DCL respectively. For RFCN, Amulet and UCF, their precision, recall and $F_\beta$ tested on adversarial samples also decrease to some degree. For sparse labeling methods (MC, LEGS and MDF), their performances against adversarial attacks is almost unchanged.}
\label{fig:attack_PRF_bar}
\end{figure}

As shown in Table~\ref{table:attack_effect}, $F_\beta$ measure of DSS and DCL drop 30\%-36\% when exposed to the adversarial samples. The adversarial attack reduces $F_\beta$ measure of RFCN, Amulet and UCF by 6\%-12\% while it only lowers $F_\beta$ of MC, LEGS, MDF by 0.7\%-1.1\%. As shown in Table~\ref{table:attack_effect}, MAE of DSS and DCL are increased by 0.176 and 0.1338 on the adversarial samples while that of RFCN, Amulet and UCF are raised by around 0.04. MAE of MC, LEGS, MDF change less than 0.01. These results indicate that DSS suffers most from the adversarial attack for the adversarial samples are synthesized using a DSS model. DCL, RFCN, Amulet and UCF are affected to different extent, which may depend on the similarity between their architectures and the pre-trained model used to launch attacks.
\begin{figure*}[!t]
\centering
\includegraphics[scale=1.0]{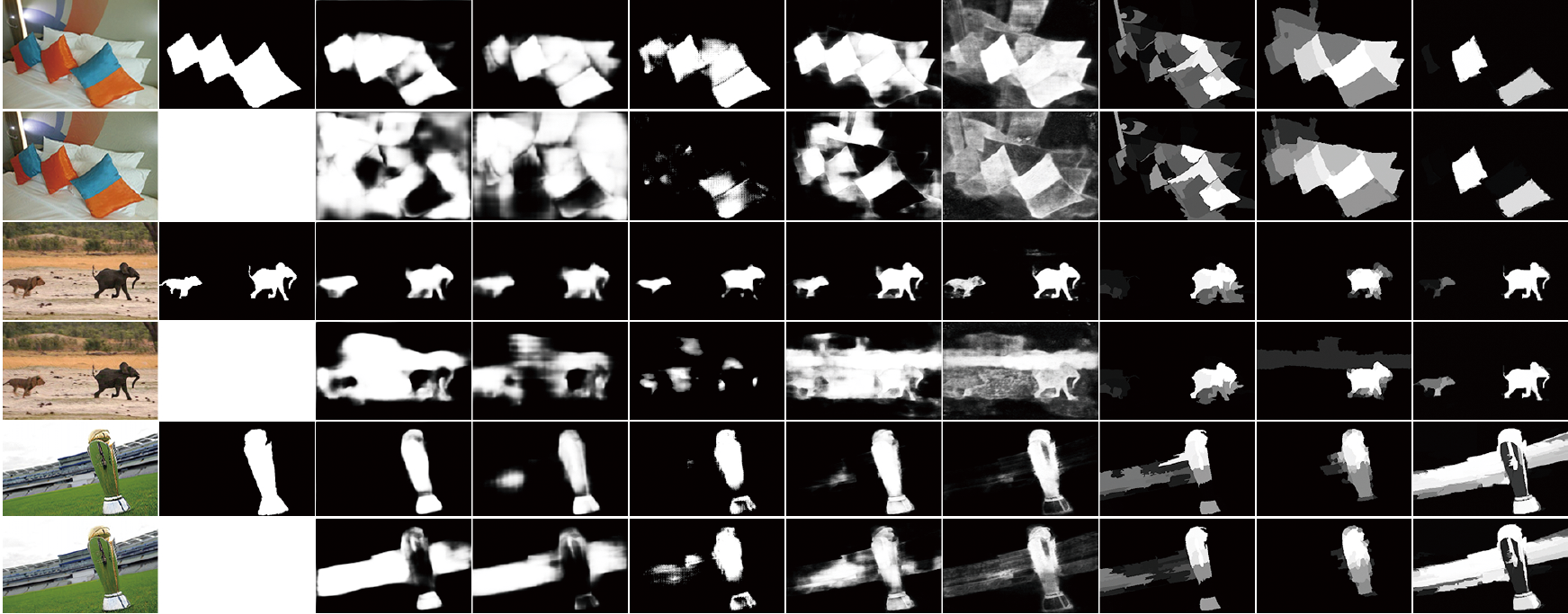}
\leftline{\scriptsize \hspace{0.35cm} Input Images \hspace{0.25cm} Ground Truth \hspace{0.5cm} DSS~\cite{HouPami18Dss} \hspace{0.7cm} DCL~\cite{LiYu16} \hspace{0.55cm} RFCN~\cite{Wang2016Saliency} \hspace{0.4cm} Amulet~\cite{zhang2017amulet} \hspace{0.5cm} UCF~\cite{zhang2017learning} \hspace{0.65cm} MC~\cite{zhao2015saliency} \hspace{0.6cm} LEGS~\cite{wang2015deep} \hspace{0.5cm} MDF~\cite{li2015visual}}
\caption{Effectiveness of Adversarial Attack. The leftmost column are input images in which the upper one is a natural image and the lower one is the corresponding adversarial samples. The second column from the left are ground truth saliency maps in which the lower position leave vacant because natural images and their adversarial samples share the same ground truth. As shown in the above examples, saliency maps predicted by FCN based salient object detection models including DSS, DCL, RFCN, Amulet, UCF are deteriorated by input images with adversarial perturbations. Segment based models such as MC, LEGS and MDF produce more consistent results between natural images and adversarial samples. }
\label{attackSample}
\end{figure*}
\begin{figure*}[!t]
\centering
\includegraphics[scale=1.0]{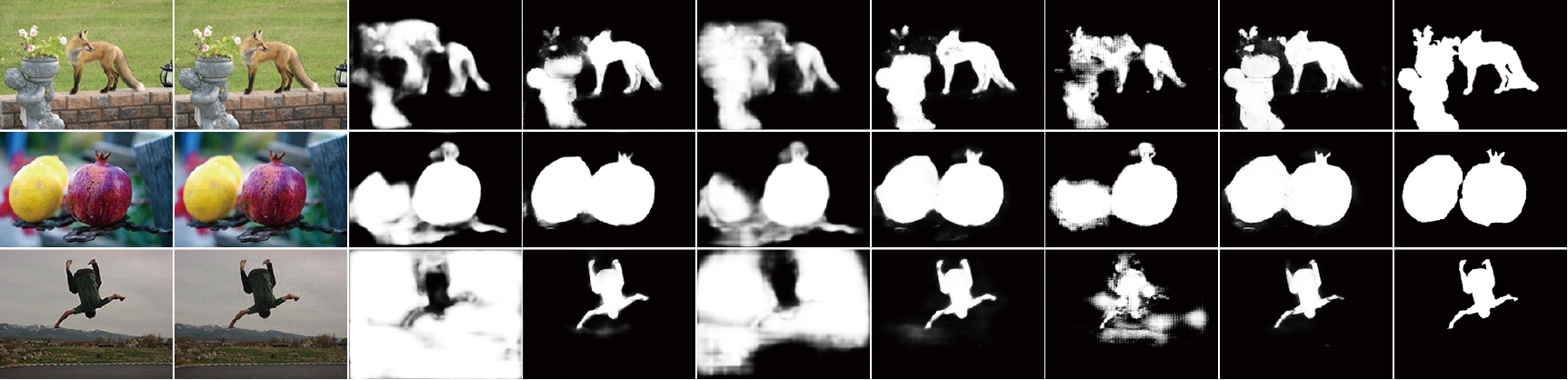}
\leftline{\scriptsize \hspace{0.35cm} Original Images\hspace{0.05cm} Adversarial Samples\hspace{0.45cm} DSS~\cite{HouPami18Dss} \hspace{0.8cm} DSS+ours \hspace{0.8cm} DCL~\cite{LiYu16} \hspace{0.75cm} DCL+ours \hspace{0.7cm} RFCN~\cite{Wang2016Saliency} \hspace{0.6cm} RFCN+ours \hspace{0.45cm} Ground Truth}
\caption{Robustness of ROSA. The leftmost column are original natural images. The second column from the left are the corresponding adversarial samples. The rightmost column are ground truth. $N$+ours denotes some neural model $N$ incorporated with our proposed method ROSA. All salient object detection models in the above are tested on adversarial samples. As shown in the above examples, our proposed framework enhances the prediction accuracy of three backbone network DSS, DCL and RFCN.}
\label{ROSAsample}
\end{figure*}

Figure~\ref{fig:attack_PR_curve} demonstrates the comparison of PR curves w.r.t these above-mentioned salient object detection models tested on natural images and adversarial samples. $N$-adv denotes some neural network $N$ tested on adversarial samples. The results tested on adversarial samples are plotted using blue solid curves while those tested on original images are draw with orange dashed curves. As shown in Figure~\ref{fig:attack_PR_curve}, the PR curves of DSS and DCL are significantly higher than DSS-adv and DCL-adv. It indicates that DSS and DCL suffer the most from adversarial samples. Because the adversarial samples are synthesized using exactly the same DSS model. DCL also severely affected by the adversarial attacks, possibly because it has similar network structure with DSS. The PR curves of RFCN-adv, Amulet-adv and UCF-adv also decline by a considerable margin respectively, in comparison to RFCN, Amulet and UCF. That is to say, adversarial samples generated by some FCN based model are transferable to degrade other FCN based methods to different extent. Even attackers are unaware of the target neural network, they still have chances to launch successful attacks using some arbitrary FCN model. It reveals that existing visual saliency models based on dense labeling are threatened by adversarial attacks. For MC, LEGS and MDF, their PR curves on natural images and adversarial samples almost completely overlap. It suggests that sparse labeling methods are quite robust against existing gradient-based attacks, since gradients propagated from different segments to the same image position very possibly conflict with each other.
\renewcommand{\topfraction}{0.85}
\newcolumntype{C}[1]{>{\centering\arraybackslash}p{#1}}
\begin{table*}[!ht]
\centering
\caption{Robustness of ROSA with DSS}
\label{table:defense_dss}
\small
\label{tab1}
\begin{tabular}{c|c C{1.0cm} C{1.3cm} C{1.3cm} C{1.6cm} C{1.4cm} C{1.4cm} C{1.4cm}}
\hline
\multirow{1}*{Dataset} & \multirow{1}*{Metric} & \multirow{1}*{DSS*} & DSS+ours*  & DSS+ours & DSS+Smooth & DSS+JPEG & DSS+Quant & DSS+TVM \\

\hline \hline
\multirow{2}*{HKU-IS-adv}    & $F_\beta$    & 0.74\% & \textcolor{green}{83.18\%} & \textcolor{red}{83.52\%} & 54.93\% & 12.61\% & 31.80\% & 77.44\% \\
                             &    MAE       & 0.7495 & \textcolor{green}{0.0654}  & \textcolor{red}{0.0644}  & 0.1831  & 0.4638  & 0.3224  & 0.0981  \\
\hline
\multirow{2}*{HKU-IS}        & $F_\beta$    & \textcolor{green}{87.50\%} &         & \textcolor{red}{88.48\%} & 85.92\% & 87.28\% & 87.07\% & 84.39\% \\
                             &    MAE       & \textcolor{green}{0.0436} &         & \textcolor{red}{0.0341}  & 0.0528  & 0.0446  & 0.0479  & 0.0670  \\
\hline
\multirow{2}*{ECSSD-adv}     & $F_\beta$    & 0.90\% & \textcolor{green}{81.48\%} & \textcolor{red}{82.26\%} & 55.09\% & 12.91\% & 34.84\% & 79.34\% \\
                             &    MAE       & 0.7763 & \textcolor{green}{0.0940}  & \textcolor{red}{0.0915}  & 0.2202  & 0.4960  & 0.3428  & 0.1242  \\
\hline
\multirow{2}*{ECSSD}         & $F_\beta$    & \textcolor{red}{87.69\%} &         & 87.33\% & 86.56\% & \textcolor{green}{87.59\%} & 87.23\% & 85.62\% \\
                             &    MAE       & \textcolor{green}{0.0608} &         & \textcolor{red}{0.0475}  & 0.0712  & 0.0618  & 0.0689  & 0.0891  \\
\hline
\multirow{2}*{DUT-OMRON-adv} & $F_\beta$    & 0.41\% & \textcolor{green}{72.27\%} & \textcolor{red}{72.31\%} & 35.41\% & 22.62\% & 26.83\% & 66.26\% \\
                             &    MAE       & 0.8038 & \textcolor{red}{0.0687}  & \textcolor{green}{0.0690}  & 0.2265  & 0.3409  & 0.2885  & 0.0896  \\
\hline
\multirow{2}*{DUT-OMRON}     & $F_\beta$    & \textcolor{green}{76.20\%}&         & \textcolor{red}{79.63\%} & 74.96\% & 76.19\% & 75.32\% & 73.70\% \\
                             &    MAE       & \textcolor{green}{0.0547} &         & \textcolor{red}{0.0469}  & 0.0591  & 0.0547  & 0.0581  & 0.0675  \\
\hline
\multirow{2}*{MSRA-B-adv}    & $F_\beta$    & 0.51\% & \textcolor{green}{86.56\%} & \textcolor{red}{86.71\%} & 49.79\% & 35.48\% & 46.35\% & 84.11\% \\
                             &    MAE       & 0.8021 & \textcolor{green}{0.0549}  & \textcolor{red}{0.0548}  & 0.2460  & 0.3376  & 0.2608  & 0.0812  \\
\hline
\multirow{2}*{MSRA-B}        & $F_\beta$    & \textcolor{green}{89.59\%}&         & \textcolor{red}{89.76\%} & 89.11\% & 89.58\% & 89.06\% & 88.63\% \\
                             &    MAE       & \textcolor{green}{0.0440} &         & \textcolor{red}{0.0366}  & 0.0484  & 0.0440  & 0.0489  & 0.0567  \\
\hline
\end{tabular}
\end{table*}
\begin{table*}[!ht]
\centering
\caption{Robustness of ROSA with DCL}
\label{table:defense_dcl}
\small
\label{tab2}
\begin{tabular}{c|c C{1.0cm} C{1.3cm} C{1.6cm} C{1.4cm} C{1.4cm} C{1.4cm}}
\hline
\multirow{1}*{Dataset} & \multirow{1}*{Metric} & \multirow{1}*{DCL} & DCL+ours & DCL+Smooth & DCL+JPEG & DCL+Quant & DCL+TVM \\

\hline \hline
\multirow{2}*{HKU-IS-adv}    & $F_\beta$    & 77.84\% & \textcolor{red}{84.38\%} & 79.89\% & 79.36\% & 79.53\% & \textcolor{green}{81.03\%} \\
                             &    MAE       & 0.0866  & \textcolor{red}{0.0695}  & 0.0843  & 0.0810  & \textcolor{green}{0.0807}  & 0.0911  \\
\hline
\multirow{2}*{HKU-IS}        & $F_\beta$    & \textcolor{green}{85.22\%} & \textcolor{red}{86.87\%} & 84.67\% & 85.08\% & 84.72\% & 83.59\% \\
                             &    MAE       & \textcolor{green}{0.0540}  & \textcolor{red}{0.0507}  & 0.0671  & 0.0550  & 0.0579  & 0.0797  \\
\hline
\multirow{2}*{ECSSD-adv}     & $F_\beta$    & 79.22\% & \textcolor{red}{83.76\%} & 80.50\% & 80.85\% & 80.58\% & \textcolor{green}{82.28\%} \\
                             &    MAE       & 0.1070  & \textcolor{red}{0.0960}  & 0.1081  & \textcolor{green}{0.1008}  & 0.1047  & 0.1229  \\
\hline
\multirow{2}*{ECSSD}         & $F_\beta$    & \textcolor{green}{85.82\%} & \textcolor{red}{86.25\%} & 85.02\% & 85.63\% & 85.41\% & 84.82\% \\
                             &    MAE       & \textcolor{green}{0.0698}  & \textcolor{red}{0.0677}  & 0.0867  & 0.0712  & 0.0783  & 0.1093  \\
\hline
\multirow{2}*{DUT-OMRON-adv} & $F_\beta$    & 60.46\% & \textcolor{red}{69.68\%} & 60.34\% & 64.46\% & 62.54\% & \textcolor{green}{67.95\%} \\
                             &    MAE       & 0.1091  & \textcolor{red}{0.0794}  & 0.1092  & 0.0934  & 0.0995  & \textcolor{green}{0.0912}  \\
\hline
\multirow{2}*{DUT-OMRON}     & $F_\beta$    & 70.30\% & \textcolor{red}{72.23\%} & 67.92\% & 70.30\% & 69.10\% & \textcolor{green}{70.66\%} \\
                             &    MAE       & \textcolor{green}{0.0723}  & \textcolor{red}{0.0683}  & 0.0891  & 0.0723  & 0.0769  & 0.0834  \\
\hline
\multirow{2}*{MSRA-B-adv}    & $F_\beta$    & 79.87\% & \textcolor{red}{87.82\%} & 83.62\% & 83.53\% & 82.68\% & \textcolor{green}{85.58\%} \\
                             &    MAE       & 0.0960  & \textcolor{red}{0.0645}  & \textcolor{green}{0.0775}  & 0.0783  & 0.0819  & 0.0837  \\
\hline
\multirow{2}*{MSRA-B}        & $F_\beta$    & 87.80\% & \textcolor{red}{89.29\%} & \textcolor{green}{87.96\%} & 87.76\% & 87.02\% & 87.19\% \\
                             &    MAE       & \textcolor{green}{0.0563}  & \textcolor{red}{0.0521}  & 0.0595  & 0.0563  & 0.0609  & 0.0743  \\
\hline
\end{tabular}
\end{table*}
Figure~\ref{fig:attack_PRF_bar} are the bar diagrams of existing visual saliency methods tested on natural images and adversarial samples. P, R, F denote precision, recall and $F_\beta$ measure respectively in the color of blue, green and yellow. The results shown in Figure~\ref{fig:attack_PRF_bar} draws similar conclusions with Figure~\ref{fig:attack_PR_curve}. Precision, recall and $F_\beta$ of DSS and DCL decrease the most seriously against adversarial attacks. For RFCN, Amulet and UCF, their precision, recall and $F_\beta$ are also harmed by adversarial samples to some degree. Segment based models such MC, LEGS and MDF are more robust and present negligible degeneration.

\subsection{Robustness of the Robust Salient Object Detection Framework}
To demonstrate the robustness of ROSA, we present extensive experiments on four datasets (HKU-IS, ECSSD, DUT-OMRON and MSRA-B), with three state-of-the-art saliency models (DSS, DCL and RFCN) as baselines. All models in the section are trained on a dataset that includes training sets of HKU-IS, DUT-OMRON and MSRAB. Adversarial samples are synthesized with a DSS model pre-trained on the above-mentioned dataset. The $L_\infty$ norm upper bound $\epsilon$ of the adversarial noise is chosen as 25. We also compare our proposed method with serveral existing defending algorithms, which are developed for robust image classification and can be transferred to other tasks. Smooth denotes a spatial smooth filter in~\cite{li2017adversarial}. JPEG~\cite{das2018shield,guo2018countering} denotes applying JPEG compression on input images before feeding them into target networks. The quality of the compressed image is set to 75 according to \cite{guo2018countering}. Quant~\cite{xu2018feature} denotes bit reduction that quantizes 8-bit RGB images into pixel values with less bits. We reduce images to 3 bits following \cite{guo2018countering}. TVM~\cite{guo2018countering} denotes total variation minimization that aims at reducing difference between adjacent pixels. TVM is implemented using \cite{Getreuer2012TV}. Table~\ref{table:defense_dss}, Table~\ref{table:defense_dcl} and Table~\ref{table:defense_rfcn} are the numeric results with a baseline model as DSS, DCL and RFCN respectively. $D$-adv denotes experiments on the adversarial samples of dataset $D$. $N$+$M$ denotes the neural network $N$ equipped with the defense method $M$.
\begin{table*}[!t]
\centering
\caption{Robustness of ROSA with RFCN}
\label{table:defense_rfcn}
\small
\label{tab3}
\begin{tabular}{c|c C{1.0cm} C{1.5cm} C{1.8cm} C{1.6cm} C{1.6cm} C{1.6cm}}
\hline
\multirow{1}*{Dataset} & \multirow{1}*{Metric} & \multirow{1}*{RFCN} &RFCN+ours &RFCN+Smooth &RFCN+JPEG &RFCN+Quant &RFCN+TVM \\

\hline
\multirow{2}*{HKU-IS-adv}    & $F_\beta$    & 76.58\% & \textcolor{red}{85.76\%} & 76.38\% & 78.36\% & \textcolor{green}{79.83\%} & 78.72\% \\
                             &    MAE       & 0.0985  & \textcolor{red}{0.0599}  & 0.1036  & 0.0916  & \textcolor{green}{0.0862}  & 0.1068  \\
\hline
\multirow{2}*{HKU-IS}        & $F_\beta$    & \textcolor{green}{86.94\%} & \textcolor{red}{87.18\%} & 85.12\% & 86.24\% & 85.65\% & 82.69\% \\
                             &    MAE       & \textcolor{red}{0.0533}  & \textcolor{green}{0.0535}  & 0.0668  & 0.0563  & 0.0612  & 0.0893  \\
\hline
\multirow{2}*{ECSSD-adv}     & $F_\beta$    & 77.24\% & \textcolor{red}{84.89\%} & 78.27\% & 79.69\% & 80.61\% & \textcolor{green}{80.80\%} \\
                             &    MAE       & 0.1290  & \textcolor{red}{0.0836}  & 0.1319  & 0.1190  & \textcolor{green}{0.1154}  & 0.1382  \\
\hline
\multirow{2}*{ECSSD}         & $F_\beta$    & \textcolor{red}{87.22\%} & 85.76\% & 85.73\% & \textcolor{green}{86.77\%} & 86.04\% & 84.05\% \\
                             &    MAE       & \textcolor{red}{0.0735}  & 0.0802  & 0.0911  & \textcolor{green}{0.0770}  & 0.0869  & 0.1207  \\
\hline
\multirow{2}*{DUT-OMRON-adv} & $F_\beta$    & 56.09\% & \textcolor{red}{72.57\%} & 60.60\% & 63.79\% & 62.30\% & \textcolor{green}{66.21\%} \\
                             &    MAE       & 0.1257  & \textcolor{red}{0.0691}  & 0.1132  & \textcolor{green}{0.1002}  & 0.1048  & 0.1029  \\
\hline
\multirow{2}*{DUT-OMRON}     & $F_\beta$    & \textcolor{green}{72.18\%} & \textcolor{red}{74.37\%} & 71.28\% & 72.15\% & 70.02\% & 70.14\% \\
                             &    MAE       & \textcolor{green}{0.0728}  & \textcolor{red}{0.0638}  & 0.0786  & 0.0728  & 0.0802  & 0.0918  \\
\hline
\multirow{2}*{MSRA-B-adv}    & $F_\beta$    & 76.32\% & \textcolor{red}{89.68\%} & 79.96\% & 83.46\% & 83.36\% & \textcolor{green}{85.14\%} \\
                             &    MAE       & 0.1210  & \textcolor{red}{0.0507}  & 0.1053  & 0.0852  & \textcolor{green}{0.0848}  & 0.0912  \\
\hline
\multirow{2}*{MSRA-B}        & $F_\beta$    & \textcolor{green}{89.45\%} & \textcolor{red}{90.53\%} & 88.43\% & 89.45\% & 88.25\% & 87.47\% \\
                             &    MAE       & \textcolor{green}{0.0518}  & \textcolor{red}{0.0469}  & 0.0605  & 0.0519  & 0.0604  & 0.0769  \\
\hline
\end{tabular}
\end{table*}
\begin{figure*}[!t]
\centering
\includegraphics[scale=0.64]{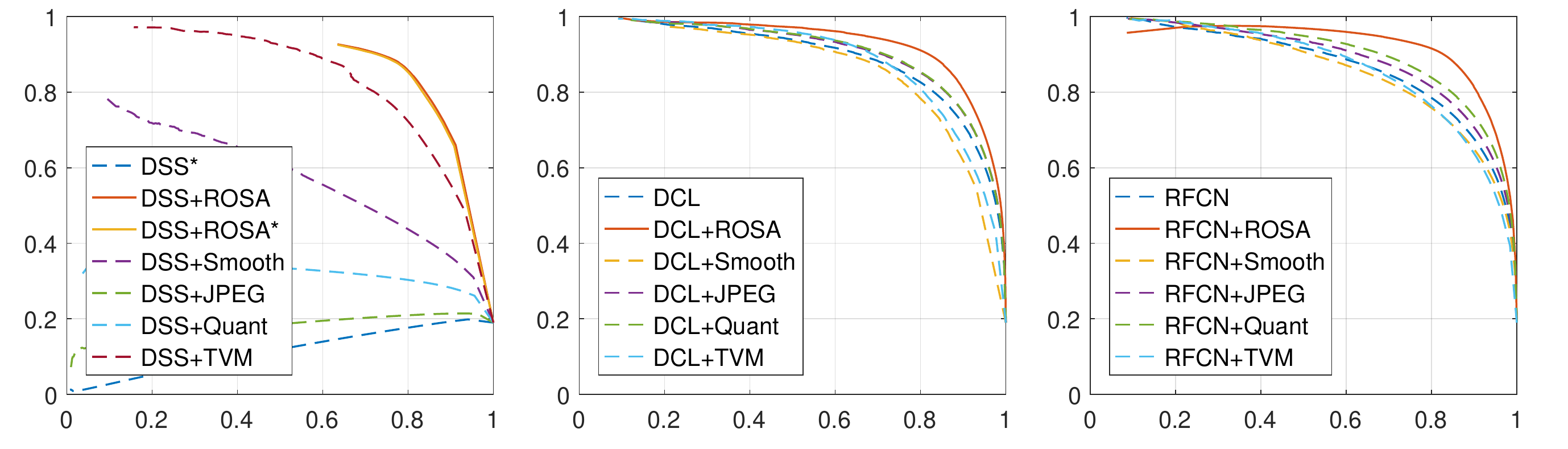}
\caption{Quantitative Analysis on the robustness of ROSA in terms of PR Curve. It shows the PR curves of existing defense methods and our proposed framework with DSS, DCL and RFCN respectively. Solid curves denote our proposed method while dashed curves denote other defense algorithms. DSS+ROSA* denotes DSS+ROSA tested on the adversarial samples synthesized by the DSS backbone of DSS+ROSA itself. As shown in the above figures, our proposed method achieves higher PR curves than other defense baselines with DSS, DCL and RFCN. }
\label{fig:PR_curve_ROSA}
\end{figure*}

As Table~\ref{table:defense_dss} shows, DSS+ours outperforms DSS* by 71.9\%-86.2\% w.r.t $F_\beta$ measure on adversarial samples. DSS* displays seriously degraded $F_\beta$ lower than 1.0\% because the adversarial samples are synthesized with the same DSS model. For fair comparison, we also attack DSS+ours with samples produced by its own DSS backbone, which is denoted as DSS+ours*. DSS+ours* still significantly surpasses DSS* by 71.86\%-86.05\% $F_\beta$. The difference of the performance between DSS+ours* and DSS+ours is quite small and less than 0.78\% $F_\beta$. Note that on natural images DSS+ours* and DSS+ours have exactly the same numerical results, since their difference lies in using DSS models of different weights to synthesize adversarial samples. For simplicity, the cell corresponding to DSS+ours tested on original images leave a blank. Compared with existing defense baselines, DSS+ours exceeds the second best DSS+TVM by 6.08\% $F_\beta$ and 0.0337 MAE on the adversarial samples of HKU-IS dataset. On ECSSD-adv dataset, our proposed method outperforms TVM by 2.92\% $F_\beta$ and 0.0327 MAE. DSS+ours surpasses the second best DSS+TVM by 6.05\% $F_\beta$ and 0.0206 MAE on DUT-OMRON-adv dataset. On the adversarial samples of MSRA-B, the proposed framework also obtains higher $F_\beta$ and smaller MAE than other defense methods. As for natural images, the performances of different models are close to each other, because of no threats caused by adversarial noises. Existing defense approaches act as small variations on input images and result in slight degeneration on performance. In most cases, the proposed framework achieves the best $F_\beta$ measure and MAE on clean input images, which suggests that our method improve the backbone model on both adversarial samples and natural images.

In the case of DCL shown in Table~\ref{table:defense_dcl}, our proposed methods presents the highest $F_\beta$ and the smallest MAE on both original images and adversarial samples of all four benchmarks. For example, DCL+ours outperforms DCL+TVM by 3.35\% $F_\beta$ and DCL+Quant by 0.0112 MAE on HKU-IS-adv. $F_\beta$ and MAE of DCL+ours are superior to those of DCL by 1.65\% and  0.0033 on HKU-IS dataset. On MSRA-B-adv data, our proposed defense framework surpasses the second best TVM by 2.24\% and Smooth by 0.013. On the natural images of MSRA-B, DCL+ours also obtains better $F_\beta$ and MAE than the second best DCL+Smooth and DCL by 1.33\% and 0.0042 respectively. In the case of RFCN shown in Table~\ref{table:defense_rfcn}, the proposed defense framework achieves the best $F_\beta$ and MAE on the adversarial samples of all four benchmarks. For example, RFCN+ours outperforms the second best RFCN+TVM by 4.09\% $F_\beta$ and RFCN+Quant by 0.0318 MAE on ECSSD-adv dataset. RFCN+ours surpasses the second best RFCN+TVM by 4.54\% and RFCN+Quant by 0.0341 MAE on MSRA-B-adv dataset. As for natural images, our proposed method RFCN+ours achieves competitive or better results than RFCN.

Figure~\ref{fig:PR_curve_ROSA} demonstrates the PR curves of existing defense baselines and the proposed method with three backbone networks, DSS, DCL and RFCN respectively. Solid curves denote the proposed defense framework while dashed ones denote existing defense algorithms. Note that in the leftmost sub-figure of Figure~\ref{fig:PR_curve_ROSA}, DSS+ROSA* denotes attacking DSS+ROSA via adversarial samples that are generated using the backbone of DSS+ROSA itself. As shown in Figure~\ref{fig:PR_curve_ROSA}, the proposed defense framework displays better PR curves than the second best TVM with DSS as backbone. On the case of DCL and RFCN, the PR curves of our proposed method are also higher than the second best Quant by a considerable margin. The existing defense algorithms perform significantly worse with DSS than DCL and RFCN, since the adversarial samples are synthesized using a DSS model. In short, our proposed framework not only significantly enhances the robustness of backbone against adversarial attacks but also demonstrates comparable or better performance on natural images. Figure~\ref{ROSAsample} presents some qualitative comparisons on the robustness of ROSA.
\begin{table}[!t]
\centering
\caption{Ablation Study of ROSA}
\label{table:abla}
\small
\begin{tabular}{ c|c c|c c }
\hline
\multirow{2}{*}{Model} &
	\multicolumn{2}{c|}{$F_\beta$ measure} &
	\multicolumn{2}{c}{MAE} \\
\cline{2-5}
         & Original & Adversarial    & Original & Adversarial \\
\hline \hline
DSS+SWS  & 82.89\% & 79.09\% & 0.0660 & 0.0802 \\
DSS+CAR  & 90.96\% & 53.86\% & 0.0428 & 0.2169 \\
DSS+ours & 85.78\% & 82.18\% & 0.0541 & 0.0683 \\
\hline
DCL+SWS & 78.19\% & 74.20\% & 0.0852 & 0.1015 \\
DCL+CAR & 88.36\% & 60.17\% & 0.0500 & 0.1754 \\
DCL+ours & 84.86\% & 81.46\% & 0.0579 & 0.0708 \\
\hline
RFCN+SWS & 77.07\% & 74.45\% & 0.0916 & 0.1004 \\
RFCN+CAR  & 88.80\% & 77.98\% & 0.0549 & 0.0904 \\
RFCN+ours & 86.05\% & 83.83\% & 0.0603 & 0.0688 \\
\hline
\end{tabular}
\end{table}
\subsection{Ablation Study}
\begin{figure}[!t]
\centering
\includegraphics[scale=0.32]{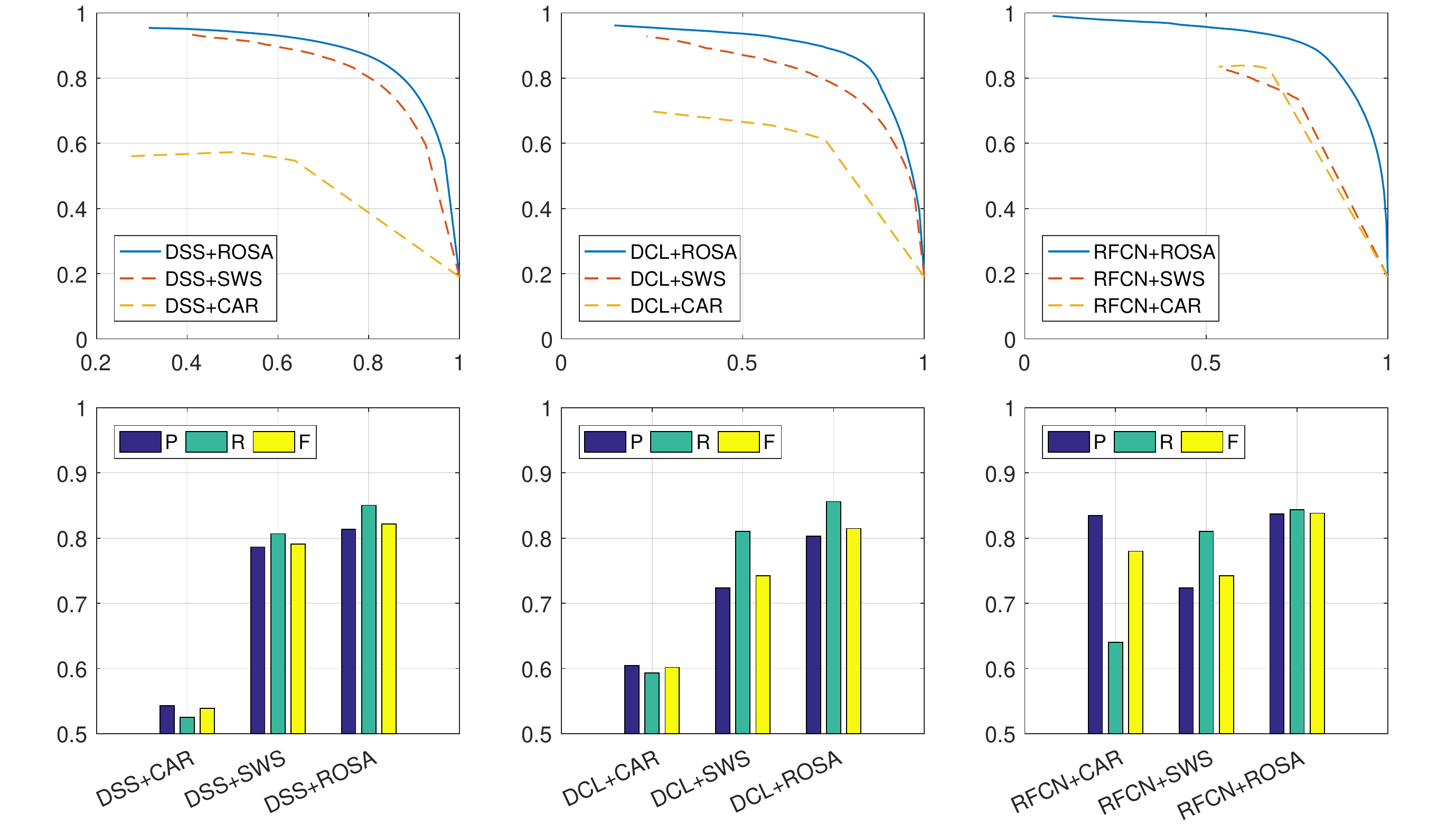}
\caption{Ablation Study of ROSA. The above compares the entire proposed framework with its separated components, segment-wise shielding component denoted as SWS, and context-aware restoration component denoted as CAR. The upper row are PR curves with DSS, DCL and RFCN while the lower one are bar diagrams of precision, recall and $F_\beta$. As shown in the above figures, the performance of the entire proposed framework is superior to those of its internal components. It suggests that the internal components in our proposed method acts as different roles to complement and enhance each other significantly.}
\label{fig:ablation}
\end{figure}
This section verifies the effectiveness of each part in the proposed ROSA framework. We integrate each component of ROSA with DSS, DCL and RFCN respectively. For simplicity, the above models are trained on MSRA-B train set and tested on HKU-IS test set. SWS denotes Segment-Wise Shielding component and CAR denotes Context-Aware Restoration component. To validate the effect of CAR/SWS, we compare *+SWS/*+CAR with *+ROSA. According to Table 3, even though *+CAR exceed *+ROSA by 5.22\%, 3.5\% and 2.75\% $F_\beta$ on natural images, *+ROSA outperforms *+CAR by 28.32\%, 21.29\% and 5.85\% against adversarial attacks, which indicates the effectiveness of SWS. MAE results in Table 3 also draw a similar conclusion. Even though MAE of *+CAR are 0.0113, 0.0079 and 0.0054 less than that of *+ROSA on natural images, *+ROSA lowers MAE by 0.1486, 0.1046 and 0.0216 on adversarial samples by a larger margin. As for components *+SWS, *+ROSA surpasses *+SWS by 2.89\%, 6.67\%, 8.98\% $F_\beta$ on original samples and 3.09\%, 7.26\%, 9.38\% $F_\beta$ on adversarial samples. Besides, *+ROSA reduces MAE by 0.0119, 0.0273, 0.0313 on natural images and 0.0119, 0.0307, 0.0316 on adversarial samples in comparison to *+SWS, which authenticates the effectiveness of CAR. We claim that SWS and CAR are two strongly coupled components. For example, DCL obtains 56.84\% $F_\beta$ against adversarial samples as shown in Table~\ref{table:attack_effect} while DCL+CAR achieves 60.17\% $F_\beta$. CAR improves DCL by 3.33\% $F_\beta$. However, DCL+ours (SWS+CAR) outperforms DCL+SWS by 7.26\% $F_\beta$ more than 3.33\%. That is to say, with SWS component, the effectiveness of CAR is more significant. The cases of DSS and RFCN draw the same conclusion. Thus SWS and CAR are not separated processing but two complementary steps of one core idea, adaptively predicting saliency for inputs with the new introduced noise to resist adversarial attacks. In short, CAR component can refine saliency maps predicted by models with SWS component. These two components are complementary and contribute to the robustness of our proposed method.
\begin{figure*}[!t]
\centering
\leftline{\scriptsize \hspace{3.6cm} Image\hspace{3.4cm} $\epsilon=10$\hspace{3.3cm} $\epsilon=20$\hspace{3.2cm} $\epsilon=30$}
\includegraphics[width=1.0\linewidth]{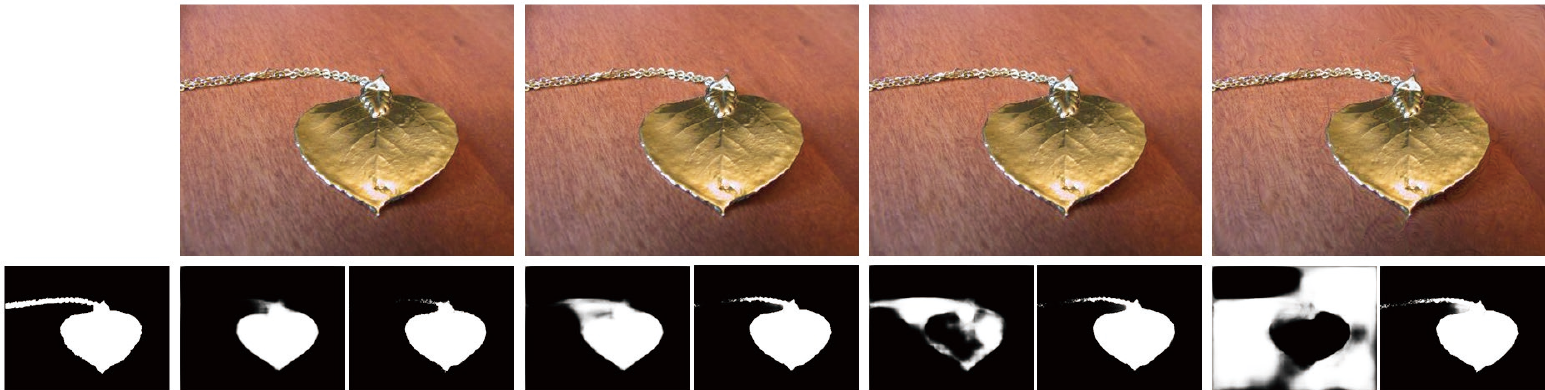}
\leftline{\scriptsize \hspace{0.16cm} Ground Truth\hspace{1.15cm} DSS\hspace{1.15cm} DSS+ours\hspace{1.2cm} DSS\hspace{1.2cm} DSS+ours\hspace{1.2cm} DSS\hspace{1.2cm} DSS+ours\hspace{1.2cm} DSS\hspace{1.2cm} DSS+ours}
\caption{Qualitative comparison among adversarial attacks of different strengths. $\epsilon$ denotes the $L_{\infty}$ norm upper bound of adversarial perturbations. As shown in the above, larger $\epsilon$ achieves stronger attack and results in worse performance of a pre-trained DSS model. But the curve-like patterns of adversarial noises are more easily observed. No matter what $\epsilon$ is set in the range of [0, 30], the proposed method denoted as DSS+ours presents stable and fine saliency maps.}
\label{fig:QualiCompAttackStrength}
\end{figure*}

Figure~\ref{fig:ablation} compares the entire proposed framework with its internal components with DSS, DCL and RFCN, on PR curves and bar diagrams of precision-recall-$F_\beta$. The upper row are PR curves while the lower one are bar diagrams. Among these PR curves, the blue solid ones denote the entire proposed method while the orange/yellow dashed ones denote the internal components SWS and CAR. In the bar diagrams, P, R and F denote precision, recall and $F_\beta$ in the color of blue, green and yellow respectively. In Figure~\ref{fig:ablation}, the entire proposed framework displays higher PR curves than its internal components with three different backbone networks. Besides, the precision, recall and $F_\beta$ of the entire proposed method also surpass those of *+SWS and *+CAR. In detail, *+CAR perform the worst with DSS and DCL, while *+SWS is close to *+CAR with RFCN. It indicates that CAR component almost cannot resist adversarial attacks without SWS component. Note that *+ROSA achieves the best and outperform *+SWS with different backbones. It suggests that CAR component can further improve *+SWS by a significant margin.
\begin{figure}[!t]
    \centering
    \subfigure[Fmeasure-Epsilon curve]{
    \begin{minipage}{0.46\linewidth}
        \begin{center}
        \includegraphics[width=1.0\linewidth]{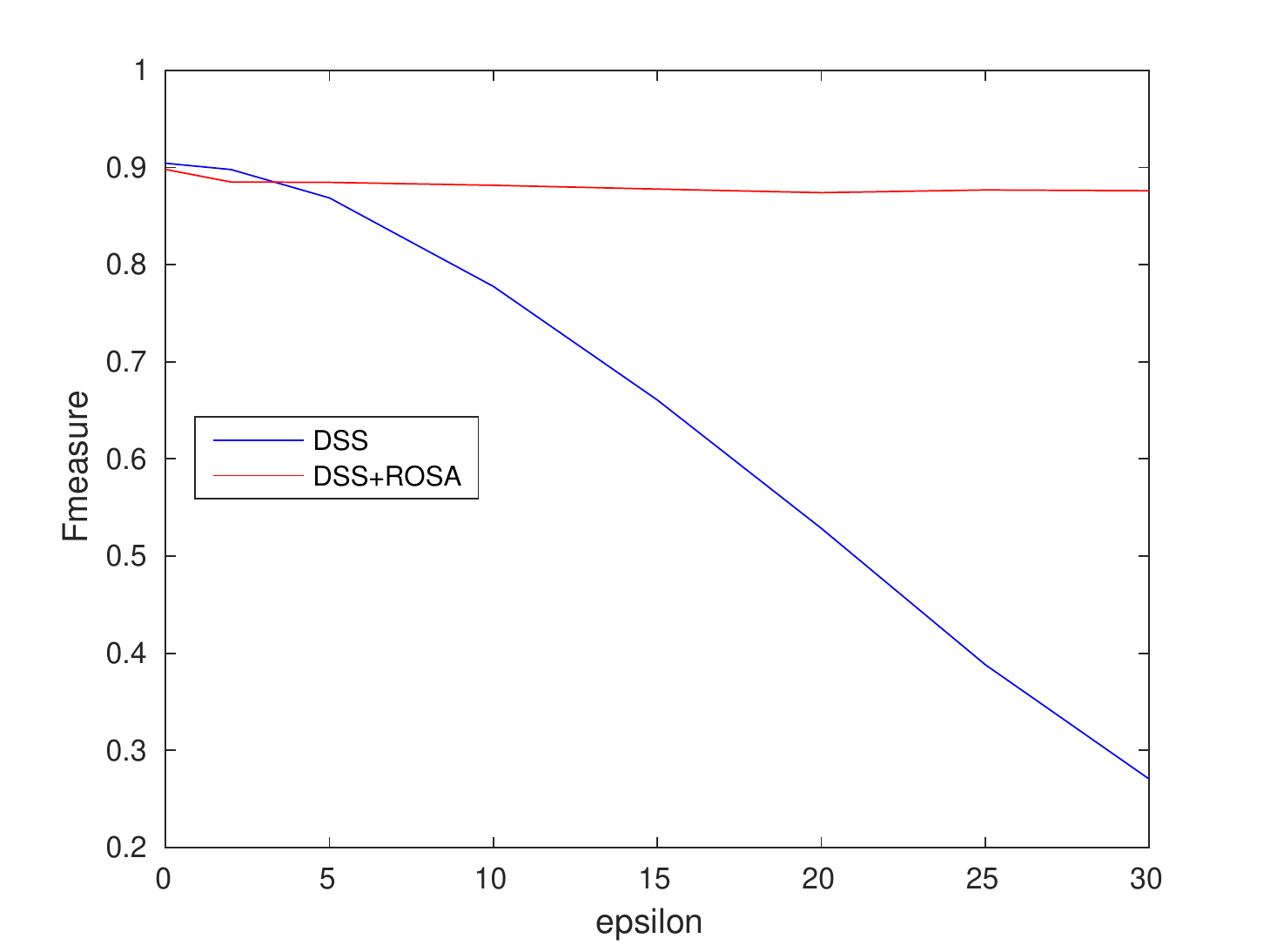}
        \end{center}
        \vspace{0.1cm}
    \end{minipage}
    }
    \subfigure[MAE-Epsilon curve]{
    \begin{minipage}{0.46\linewidth}
        \begin{center}
        \includegraphics[width=1.01\linewidth]{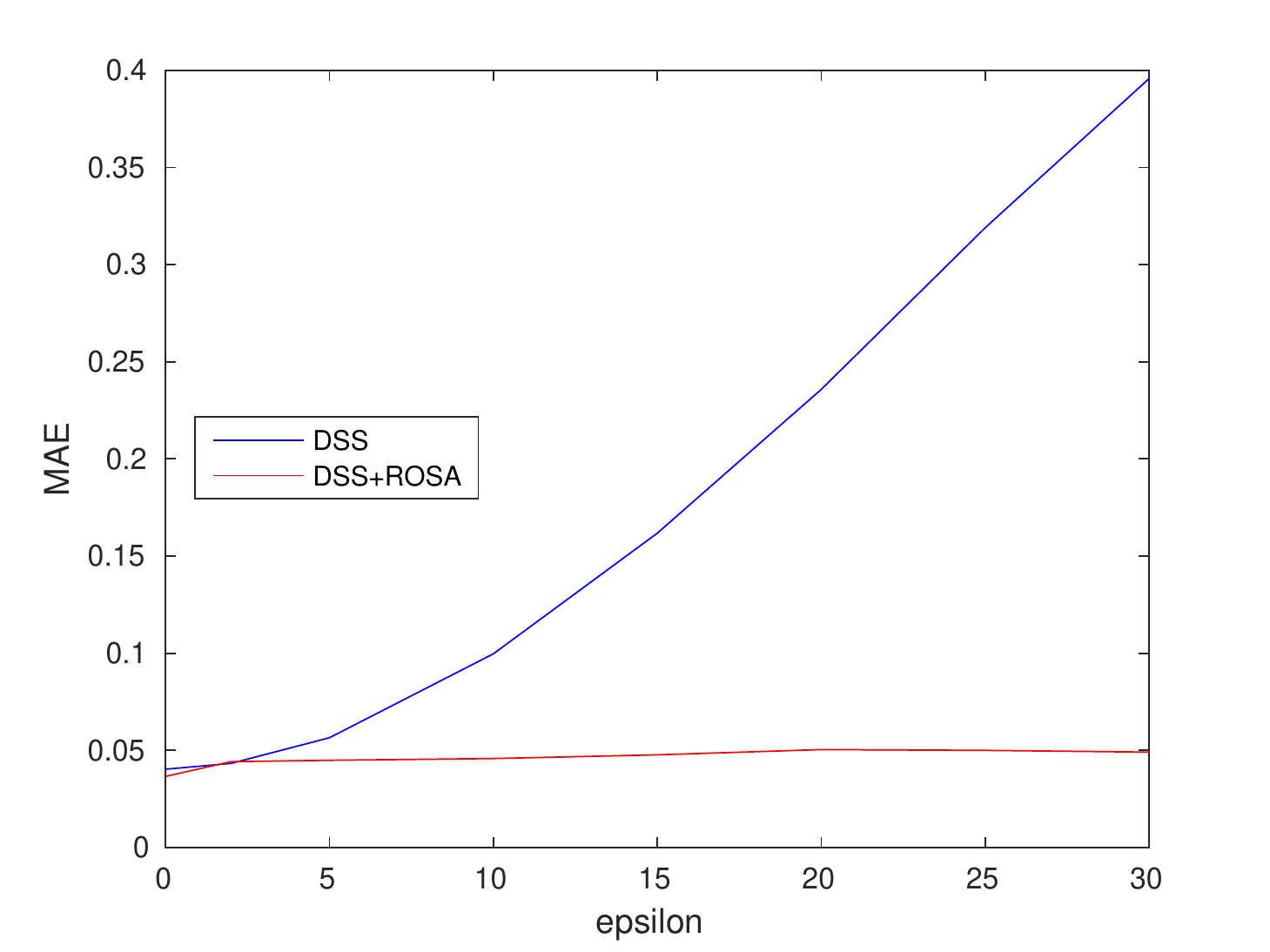}
        \end{center}
        \vspace{0.1cm}
    \end{minipage}
    }
    \caption{Investigation of Attack Strength. Epsilon denotes the $L_{\infty}$ norm upper bound of adversarial perturbations. Fmeasure denotes the evaluation metric, $F_\beta$-measure.}
    \label{fig:attackStrength}
\end{figure}
\subsection{Investigation of Attack Strength}
In this section, we investigate how a hyper-parameter, the upper bound of $L_{\infty}$ norm (denoted as $\epsilon$), affects the strength of the proposed adversarial attack. We also study how our proposed defense method performs against adversarial attacks of different strengths. For efficiency, 500 images are randomly selected from the test set of MSRA-B dataset. These 500 images are named MSRA-B500. A list of $\epsilon$ is sampled in the range of [0, 30]. For each $\epsilon$, a set of adversarial samples is synthesized for the whole MSRA-B500. These adversarial samples are computed using some pre-trained DSS model as target network. Each set of adversarial samples is tested by another trained DSS model, and our proposed method with DSS respectively. $F_\beta$-measure and MAE are calculated for each set of adversarial samples. We plot these results as Fmeasure-Epsilon curves and MAE-Epsilon curves in Figure~\ref{fig:attackStrength}. As Figure~\ref{fig:attackStrength} shows, the blue curve denotes the performance of DSS while the red one represents the proposed method (denoted as DSS+ROSA). As $\epsilon$ increases, the $F_\beta$ of DSS drops dramatically. It suggests that the strength of an adversarial attack grows with the increase of its $L_{\infty}$ norm upper bound. Notice that as $\epsilon$ rises, the performance of our proposed method only degrades slightly and then becomes stable. It indicates that the proposed defense framework is robust to adversarial samples of different strengths. Figure~\ref{fig:QualiCompAttackStrength} demonstrates a qualitative comparison among adversarial attacks of different strengths. Setting $\epsilon$ as 30 achieves the strongest attack and DSS incorrectly predicts the reverse of the ground truth as salient regions. However, the adversarial noise is perceptible and curve-like patterns can be observed in the top right of the adversarial sample. For $\epsilon = 20$, the adversarial perturbations are hard to spot and the DSS model is still seriously affected. Thus we suggest that choosing $\epsilon$ around 20 helps launch a strong and quasi-imperceptible adversarial attack on salient object detection models.
\section{Conclusion}
In this paper we for the first time achieve successful attacks on state-of-the-art visual saliency methods. We experimentally confirm that existing FCN-based models are sensitive to adversarial perturbation. In addition, this paper proposes a novel salient object detection framework that
first brings some new generic noise to input images, and then adaptively detects salient objects for the inputs with the new noise. The proposed framework is instantiated by an arbitrary FCN based backbone network, a segment-wise shielding component and a context-aware restoration component. Experimental results suggest that these two components are strongly coupled and significantly complement each other. Besides, extensive comparisons show that the entire framework can effectively strengthen the robustness of FCN-based saliency models, superior to existing defense baselines. We believe that developing an accurate, fast and robust model will be a new trend in salient object detection.

\ifCLASSOPTIONcaptionsoff
  \newpage
\fi



\bibliographystyle{IEEEtran}
\bibliography{paper}

\begin{thebibliography}{10}
\providecommand{\url}[1]{#1}
\csname url@samestyle\endcsname
\providecommand{\newblock}{\relax}
\providecommand{\bibinfo}[2]{#2}
\providecommand{\BIBentrySTDinterwordspacing}{\spaceskip=0pt\relax}
\providecommand{\BIBentryALTinterwordstretchfactor}{4}
\providecommand{\BIBentryALTinterwordspacing}{\spaceskip=\fontdimen2\font plus
\BIBentryALTinterwordstretchfactor\fontdimen3\font minus
  \fontdimen4\font\relax}
\providecommand{\BIBforeignlanguage}[2]{{%
\expandafter\ifx\csname l@#1\endcsname\relax
\typeout{** WARNING: IEEEtran.bst: No hyphenation pattern has been}%
\typeout{** loaded for the language `#1'. Using the pattern for}%
\typeout{** the default language instead.}%
\else
\language=\csname l@#1\endcsname
\fi
#2}}
\providecommand{\BIBdecl}{\relax}
\BIBdecl

\bibitem{yu2010object}
Y.~Yu, G.~K. Mann, and R.~G. Gosine, ``An object-based visual attention model
  for robotic applications,'' \emph{IEEE Transactions on Systems, Man, and
  Cybernetics, Part B (Cybernetics)}, vol.~40, no.~5, pp. 1398--1412, 2010.

\bibitem{goferman2012context}
S.~Goferman, L.~Zelnik-Manor, and A.~Tal, ``Context-aware saliency detection,''
  \emph{IEEE Transactions on Pattern Analysis and Machine Intelligence},
  vol.~34, no.~10, pp. 1915--1926, 2012.

\bibitem{wei2017stc}
Y.~Wei, X.~Liang, Y.~Chen, X.~Shen, M.-M. Cheng, J.~Feng, Y.~Zhao, and S.~Yan,
  ``Stc: A simple to complex framework for weakly-supervised semantic
  segmentation,'' \emph{IEEE Transactions on Pattern Analysis and Machine
  Intelligence}, vol.~39, no.~11, pp. 2314--2320, 2017.

\bibitem{Wang2014VOC}
\BIBentryALTinterwordspacing
C.~Wang, Y.~Guo, J.~Zhu, L.~Wang, and W.~Wang, ``Video object co-segmentation
  via subspace clustering and quadratic pseudo-boolean optimization in an mrf
  framework,'' \emph{Trans. Multi.}, vol.~16, no.~4, pp. 903--916, Jun. 2014.
  [Online]. Available: \url{https://doi.org/10.1109/TMM.2014.2306393}
\BIBentrySTDinterwordspacing

\bibitem{bi2014person}
S.~Bi, G.~Li, and Y.~Yu, ``Person re-identification using multiple experts with
  random subspaces,'' \emph{Journal of Image and Graphics}, vol.~2, no.~2,
  2014.

\bibitem{HouPami18Dss}
Q.~Hou, M.-M. Cheng, X.~Hu, A.~Borji, Z.~Tu, and P.~Torr, ``Deeply supervised
  salient object detection with short connections,'' \emph{IEEE Transactions on
  Pattern Analysis and Machine Intelligence}, 2018.

\bibitem{deng2009imagenet}
J.~Deng, W.~Dong, R.~Socher, L.-J. Li, K.~Li, and L.~Fei-Fei, ``Imagenet: A
  large-scale hierarchical image database,'' in \emph{Proceedings of the IEEE
  Conference on Computer Vision and Pattern Recognition}, 2009, pp. 248--255.

\bibitem{cheng2013efficient}
M.-M. Cheng, J.~Warrell, W.-Y. Lin, S.~Zheng, V.~Vineet, and N.~Crook,
  ``Efficient salient region detection with soft image abstraction,'' in
  \emph{Proceedings of International Conference on Computer Vision}, 2013, pp.
  1529--1536.

\bibitem{yang2013saliency}
C.~Yang, L.~Zhang, H.~Lu, X.~Ruan, and M.-H. Yang, ``Saliency detection via
  graph-based manifold ranking,'' in \emph{Proceedings of the IEEE Conference
  on Computer Vision and Pattern Recognition}, 2013, pp. 3166--3173.

\bibitem{jiang2013salient}
H.~Jiang, J.~Wang, Z.~Yuan, Y.~Wu, N.~Zheng, and S.~Li, ``Salient object
  detection: A discriminative regional feature integration approach,'' in
  \emph{Proceedings of the IEEE Conference on Computer Vision and Pattern
  Recognition}, 2013, pp. 2083--2090.

\bibitem{cheng2015global}
M.-M. Cheng, N.~J. Mitra, X.~Huang, P.~H. Torr, and S.-M. Hu, ``Global contrast
  based salient region detection,'' \emph{IEEE Transactions on Pattern Analysis
  and Machine Intelligence}, vol.~37, no.~3, pp. 569--582, 2015.

\bibitem{huang2017300}
X.~Huang and Y.-J. Zhang, ``300-fps salient object detection via minimum
  directional contrast,'' \emph{IEEE Transactions on Image Processing},
  vol.~26, no.~9, pp. 4243--4254, 2017.

\bibitem{tu2016real}
W.-C. Tu, S.~He, Q.~Yang, and S.-Y. Chien, ``Real-time salient object detection
  with a minimum spanning tree,'' in \emph{Proceedings of International
  Conference on Computer Vision}, 2016, pp. 2334--2342.

\bibitem{zhang2015minimum}
J.~Zhang, S.~Sclaroff, Z.~Lin, X.~Shen, B.~Price, and R.~Mech, ``Minimum
  barrier salient object detection at 80 fps,'' in \emph{Proceedings of
  International Conference on Computer Vision}, 2015, pp. 1404--1412.

\bibitem{zhu2014saliency}
W.~Zhu, S.~Liang, Y.~Wei, and J.~Sun, ``Saliency optimization from robust
  background detection,'' in \emph{Proceedings of the IEEE Conference on
  Computer Vision and Pattern Recognition}, 2014, pp. 2814--2821.

\bibitem{zhang2017ranking}
L.~Zhang, C.~Yang, H.~Lu, X.~Ruan, and M.-H. Yang, ``Ranking saliency,''
  \emph{IEEE transactions on pattern analysis and machine intelligence},
  vol.~39, no.~9, pp. 1892--1904, 2017.

\bibitem{zhang2018saliency}
L.~Zhang, J.~Ai, B.~Jiang, H.~Lu, and X.~Li, ``Saliency detection via absorbing
  markov chain with learnt transition probability,'' \emph{IEEE Transactions on
  Image Processing}, vol.~27, no.~2, pp. 987--998, 2018.

\bibitem{li2015visual}
G.~Li and Y.~Yu, ``Visual saliency based on multiscale deep features,'' in
  \emph{Proceedings of the IEEE Conference on Computer Vision and Pattern
  Recognition}, 2015, pp. 5455--5463.

\bibitem{kim2016shape}
J.~Kim and V.~Pavlovic, ``A shape-based approach for salient object detection
  using deep learning,'' in \emph{Proceedings of European Conference on
  Computer Vision}.\hskip 1em plus 0.5em minus 0.4em\relax Springer, 2016, pp.
  455--470.

\bibitem{li-mdf}
G.~Li and Y.~Yu, ``Visual saliency detection based on multiscale deep cnn
  features,'' \emph{IEEE Transactions on Image Processing}, vol.~25, no.~11,
  pp. 5012--5024, 2016.

\bibitem{wang2015deep}
L.~Wang, H.~Lu, X.~Ruan, and M.-H. Yang, ``Deep networks for saliency detection
  via local estimation and global search,'' in \emph{Proceedings of
  International Conference on Computer Vision}, 2015, pp. 3183--3192.

\bibitem{zhao2015saliency}
R.~Zhao, W.~Ouyang, H.~Li, and X.~Wang, ``Saliency detection by multi-context
  deep learning,'' in \emph{Proceedings of the IEEE Conference on Computer
  Vision and Pattern Recognition}, 2015, pp. 1265--1274.

\bibitem{qin2018hierarchical}
Y.~Qin, M.~Feng, H.~Lu, and G.~W. Cottrell, ``Hierarchical cellular automata
  for visual saliency,'' \emph{International Journal of Computer Vision}, vol.
  126, no.~7, pp. 751--770, 2018.

\bibitem{li-dcl}
G.~Li and Y.~Yu, ``Contrast-oriented deep neural networks for salient object
  detection,'' \emph{IEEE Transactions on Neural Networks and Learning
  Systems}, vol.~29, no.~12, pp. 6038--6051, 2018.

\bibitem{li-instance}
G.~Li, Y.~Xie, L.~Lin, and Y.~Yu, ``Instance-level salient object
  segmentation,'' in \emph{Proceedings of the IEEE conference on computer
  vision and pattern recognition (CVPR)}, 2017, pp. 2386--2395.

\bibitem{liweaksal}
G.~Li, Y.~Xie, and L.~Lin, ``Weakly supervised salient object detection using
  image labels,'' in \emph{Proceedings of the Thirty-Second AAAI Conference on
  Artificial Intelligence (AAAI)}, 2018.

\bibitem{luo2017non}
Z.~Luo, A.~K. Mishra, A.~Achkar, J.~A. Eichel, S.~Li, and P.-M. Jodoin,
  ``Non-local deep features for salient object detection.'' in
  \emph{Proceedings of the IEEE Conference on Computer Vision and Pattern
  Recognition}, vol.~2, no.~6, 2017, p.~7.

\bibitem{zhang2017amulet}
P.~Zhang, D.~Wang, H.~Lu, H.~Wang, and X.~Ruan, ``Amulet: Aggregating
  multi-level convolutional features for salient object detection,'' in
  \emph{Proceedings of International Conference on Computer Vision}, 2017, pp.
  202--211.

\bibitem{zhang2017learning}
P.~Zhang, D.~Wang, H.~Lu, H.~Wang, and B.~Yin, ``Learning uncertain
  convolutional features for accurate saliency detection,'' in
  \emph{Proceedings of International Conference on Computer Vision}, 2017, pp.
  212--221.

\bibitem{wang2018detect}
T.~Wang, L.~Zhang, S.~Wang, H.~Lu, G.~Yang, X.~Ruan, and A.~Borji, ``Detect
  globally, refine locally: A novel approach to saliency detection,'' in
  \emph{Proceedings of the IEEE Conference on Computer Vision and Pattern
  Recognition}, 2018, pp. 3127--3135.

\bibitem{chen2018reverse}
S.~Chen, X.~Tan, B.~Wang, and X.~Hu, ``Reverse attention for salient object
  detection,'' \emph{Proceedings of European Conference on Computer Vision},
  pp. 234--250, 2018.

\bibitem{Chen2018Emb}
S.~Chen, B.~Wang, X.~Tan, and X.~Hu, ``Embedding attention and residual network
  for accurate salient object detection,'' \emph{IEEE Transactions on
  Cybernetics}, pp. 1--13, 2018.

\bibitem{wang2018salient}
L.~Wang, L.~Wang, H.~Lu, P.~Zhang, and X.~Ruan, ``Salient object detection with
  recurrent fully convolutional networks,'' \emph{IEEE Transactions on Pattern
  Analysis and Machine Intelligence}, 2018.

\bibitem{LiYu16}
G.~Li and Y.~Yu, ``Deep contrast learning for salient object detection,'' in
  \emph{Proceedings of the IEEE Conference on Computer Vision and Pattern
  Recognition}, June 2016, pp. 478--487.

\bibitem{Wang2016Saliency}
L.~Wang, L.~Wang, H.~Lu, P.~Zhang, and R.~Xiang, ``Saliency detection with
  recurrent fully convolutional networks,'' in \emph{Proceedings of the
  European Conference on Computer Vision}, 2016, pp. 825--841.

\bibitem{xie2015holistically}
S.~Xie and Z.~Tu, ``Holistically-nested edge detection,'' in \emph{Proceedings
  of International Conference on Computer Vision}, 2015, pp. 1395--1403.

\bibitem{Goodfellow2015Explaining}
I.~J. Goodfellow, J.~Shlens, and C.~Szegedy, ``Explaining and harnessing
  adversarial examples,'' in \emph{Proceedings of International Conference on
  Learning Representations}, 2015.

\bibitem{Dong_2018_CVPR}
Y.~Dong, F.~Liao, T.~Pang, H.~Su, J.~Zhu, X.~Hu, and J.~Li, ``Boosting
  adversarial attacks with momentum,'' in \emph{Proceedings of the IEEE
  Conference on Computer Vision and Pattern Recognition}, June 2018.

\bibitem{moosavi2016deepfool}
S.-M. Moosavi-Dezfooli, A.~Fawzi, and P.~Frossard, ``Deepfool: a simple and
  accurate method to fool deep neural networks,'' in \emph{Proceedings of the
  IEEE Conference on Computer Vision and Pattern Recognition}, 2016, pp.
  2574--2582.

\bibitem{NIPS2017_7273}
\BIBentryALTinterwordspacing
M.~M. Cisse, Y.~Adi, N.~Neverova, and J.~Keshet, ``Houdini: Fooling deep
  structured visual and speech recognition models with adversarial examples,''
  in \emph{Advances in Neural Information Processing Systems}, I.~Guyon, U.~V.
  Luxburg, S.~Bengio, H.~Wallach, R.~Fergus, S.~Vishwanathan, and R.~Garnett,
  Eds.\hskip 1em plus 0.5em minus 0.4em\relax Curran Associates, Inc., 2017,
  pp. 6977--6987. [Online]. Available:
  \url{http://papers.nips.cc/paper/7273-houdini-fooling-deep-structured-visual-and-speech-recognition-models-with-adversarial-examples.pdf}
\BIBentrySTDinterwordspacing

\bibitem{kurakin2017adversarial}
A.~Kurakin, I.~J. Goodfellow, and S.~Bengio, ``Adversarial machine learning at
  scale,'' \emph{Proceedings of International Conference on Learning
  Representations}, 2017.

\bibitem{liu2017delving}
Y.~Liu, X.~Chen, C.~Liu, and D.~Song, ``Delving into transferable adversarial
  examples and black-box attacks,'' \emph{Proceedings of International
  Conference on Learning Representations}, 2017.

\bibitem{Moosavi-Dezfooli_2017_CVPR}
S.-M. Moosavi-Dezfooli, A.~Fawzi, O.~Fawzi, and P.~Frossard, ``Universal
  adversarial perturbations,'' in \emph{Proceedings of the IEEE Conference on
  Computer Vision and Pattern Recognition}, July 2017.

\bibitem{madry2018towards}
A.~Madry, A.~Makelov, L.~Schmidt, D.~Tsipras, and A.~Vladu, ``Towards deep
  learning models resistant to adversarial attacks,'' \emph{Proceedings of
  International Conference on Learning Representations}, 2018.

\bibitem{nguyen2015deep}
A.~Nguyen, J.~Yosinski, and J.~Clune, ``Deep neural networks are easily fooled:
  High confidence predictions for unrecognizable images,'' in \emph{Proceedings
  of the IEEE Conference on Computer Vision and Pattern Recognition}, 2015, pp.
  427--436.

\bibitem{xiao2018spatially}
C.~Xiao, J.~Zhu, B.~Li, W.~He, M.~Liu, and D.~Song, ``Spatially transformed
  adversarial examples,'' \emph{Proceedings of International Conference on
  Learning Representations}, 2018.

\bibitem{Poursaeed_2018_CVPR}
O.~Poursaeed, I.~Katsman, B.~Gao, and S.~Belongie, ``Generative adversarial
  perturbations,'' in \emph{Proceedings of the IEEE Conference on Computer
  Vision and Pattern Recognition}, June 2018.

\bibitem{zhao2018generating}
Z.~Zhao, D.~Dua, and S.~Singh, ``Generating natural adversarial examples,''
  \emph{Proceedings of International Conference on Learning Representations},
  2018.

\bibitem{szegedy2014intriguing}
C.~Szegedy, W.~Zaremba, I.~Sutskever, J.~Bruna, D.~Erhan, I.~Goodfellow, and
  R.~Fergus, ``Intriguing properties of neural networks,'' in \emph{Proceedings
  of International Conference on Learning Representations}, 2014.

\bibitem{pmlr-v80-dai18b}
H.~Dai, H.~Li, T.~Tian, X.~Huang, L.~Wang, J.~Zhu, and L.~Song, ``Adversarial
  attack on graph structured data,'' in \emph{Proceedings of the 35th
  International Conference on Machine Learning}, ser. Proceedings of Machine
  Learning Research, J.~Dy and A.~Krause, Eds., vol.~80.\hskip 1em plus 0.5em
  minus 0.4em\relax Stockholmsm?ssan, Stockholm Sweden: PMLR, 10--15 Jul 2018,
  pp. 1115--1124.

\bibitem{papernot2016distillation}
N.~Papernot, P.~McDaniel, X.~Wu, S.~Jha, and A.~Swami, ``Distillation as a
  defense to adversarial perturbations against deep neural networks,'' in
  \emph{2016 IEEE Symposium on Security and Privacy (SP)}.\hskip 1em plus 0.5em
  minus 0.4em\relax IEEE, 2016, pp. 582--597.

\bibitem{metzen2017on}
J.~H. Metzen, T.~Genewein, V.~Fischer, and B.~Bischoff, ``On detecting
  adversarial perturbations,'' \emph{Proceedings of International Conference on
  Learning Representations}, 2017.

\bibitem{lu2017safetynet}
J.~Lu, T.~Issaranon, and D.~A. Forsyth, ``Safetynet: Detecting and rejecting
  adversarial examples robustly.'' in \emph{Proceedings of International
  Conference on Computer Vision}, 2017, pp. 446--454.

\bibitem{guo2018countering}
\BIBentryALTinterwordspacing
C.~Guo, M.~Rana, M.~Cisse, and L.~van~der Maaten, ``Countering adversarial
  images using input transformations,'' in \emph{Proceedings of International
  Conference on Learning Representations}, 2018. [Online]. Available:
  \url{https://openreview.net/forum?id=SyJ7ClWCb}
\BIBentrySTDinterwordspacing

\bibitem{xie2018mitigating}
C.~Xie, J.~Wang, Z.~Zhang, Z.~Ren, and A.~L. Yuille, ``Mitigating adversarial
  effects through randomization,'' \emph{Proceedings of International
  Conference on Learning Representations}, 2018.

\bibitem{song2018pixeldefend}
Y.~Song, T.~Kim, S.~Nowozin, S.~Ermon, and N.~Kushman, ``Pixeldefend:
  Leveraging generative models to understand and defend against adversarial
  examples,'' \emph{Proceedings of International Conference on Learning
  Representations}, 2018.

\bibitem{Liao_2018_CVPR}
F.~Liao, M.~Liang, Y.~Dong, T.~Pang, X.~Hu, and J.~Zhu, ``Defense against
  adversarial attacks using high-level representation guided denoiser,'' in
  \emph{Proceedings of the IEEE Conference on Computer Vision and Pattern
  Recognition}, June 2018.

\bibitem{Akhtar_2018_CVPR}
N.~Akhtar, J.~Liu, and A.~Mian, ``Defense against universal adversarial
  perturbations,'' in \emph{Proceedings of the IEEE Conference on Computer
  Vision and Pattern Recognition}, June 2018.

\bibitem{Wang2017VVV}
\BIBentryALTinterwordspacing
C.~Wang, J.~Zhu, Y.~Guo, and W.~Wang, ``Video vectorization via tetrahedral
  remeshing,'' \emph{Transaction on Image Processing}, vol.~26, no.~4, pp.
  1833--1844, Apr. 2017. [Online]. Available:
  \url{https://doi.org/10.1109/TIP.2017.2666742}
\BIBentrySTDinterwordspacing

\bibitem{he2016deep}
K.~He, X.~Zhang, S.~Ren, and J.~Sun, ``Deep residual learning for image
  recognition,'' in \emph{Proceedings of the IEEE Conference on Computer Vision
  and Pattern Recognition}, 2016, pp. 770--778.

\bibitem{simonyan2014very}
K.~Simonyan and A.~Zisserman, ``Very deep convolutional networks for
  large-scale image recognition,'' \emph{arXiv preprint arXiv:1409.1556}, 2014.

\bibitem{xie2017adversarial}
C.~Xie, J.~Wang, Z.~Zhang, Y.~Zhou, L.~Xie, and A.~Yuille, ``Adversarial
  examples for semantic segmentation and object detection,'' in
  \emph{Proceedings of International Conference on Computer Vision}, 2017.

\bibitem{Achanta2009Frequency}
R.~Achanta, S.~Hemami, F.~Estrada, and S.~Susstrunk, ``Frequency-tuned salient
  region detection,'' in \emph{Proceedings of the IEEE Conference on Computer
  Vision and Pattern Recognition}, 2009, pp. 1597--1604.

\bibitem{krahenbuhl2011efficient}
P.~Kr{\"a}henb{\"u}hl and V.~Koltun, ``Efficient inference in fully connected
  crfs with gaussian edge potentials,'' in \emph{Proceedings of Advances in
  Neural Information Processing Systems}, 2011, pp. 109--117.

\bibitem{zheng2015conditional}
S.~Zheng, S.~Jayasumana, B.~Romeraparedes, V.~Vineet, Z.~Su, D.~Du, C.~Huang,
  and P.~H.~S. Torr, ``Conditional random fields as recurrent neural
  networks,'' \emph{Proceedings of International Conference on Computer
  Vision}, pp. 1529--1537, 2015.

\bibitem{Liu2007Learning}
T.~Liu, J.~Sun, N.~N. Zheng, X.~Tang, and H.~Y. Shum, ``Learning to detect a
  salient object,'' in \emph{Proceedings of the IEEE Conference on Computer
  Vision and Pattern Recognition}, 2007, pp. 1--8.

\bibitem{shi2016hierarchical}
J.~Shi, Q.~Yan, L.~Xu, and J.~Jia, ``Hierarchical image saliency detection on
  extended cssd,'' \emph{IEEE Transactions on Pattern Analysis and Mchine
  Intelligence}, vol.~38, no.~4, pp. 717--729, 2016.

\bibitem{li2017adversarial}
X.~Li and F.~Li, ``Adversarial examples detection in deep networks with
  convolutional filter statistics.'' in \emph{Proceedings of International
  Conference on Computer Vision}, 2017, pp. 5775--5783.

\bibitem{das2018shield}
N.~Das, M.~Shanbhogue, S.~Chen, F.~Hohman, S.~Li, L.~Chen, M.~E. Kounavis, and
  D.~H. Chau, ``Shield: Fast, practical defense and vaccination for deep
  learning using jpeg compression,'' \emph{knowledge discovery and data
  mining}, pp. 196--204, 2018.

\bibitem{xu2018feature}
W.~Xu, D.~Evans, and Y.~Qi, ``Feature squeezing: Detecting adversarial examples
  in deep neural networks.'' \emph{network and distributed system security
  symposium}, 2018.

\bibitem{Getreuer2012TV}
P.~Getreuer, ``Rudin-osher-fatemi total variation denoising using split
  bregman,'' \emph{Image Processing On Line}, vol.~2, pp. 74--95, 05 2012.

\end{thebibliography}

\vspace{-10mm}
\begin{IEEEbiography}[{\includegraphics[width=1in,height=1.25in,clip,keepaspectratio]{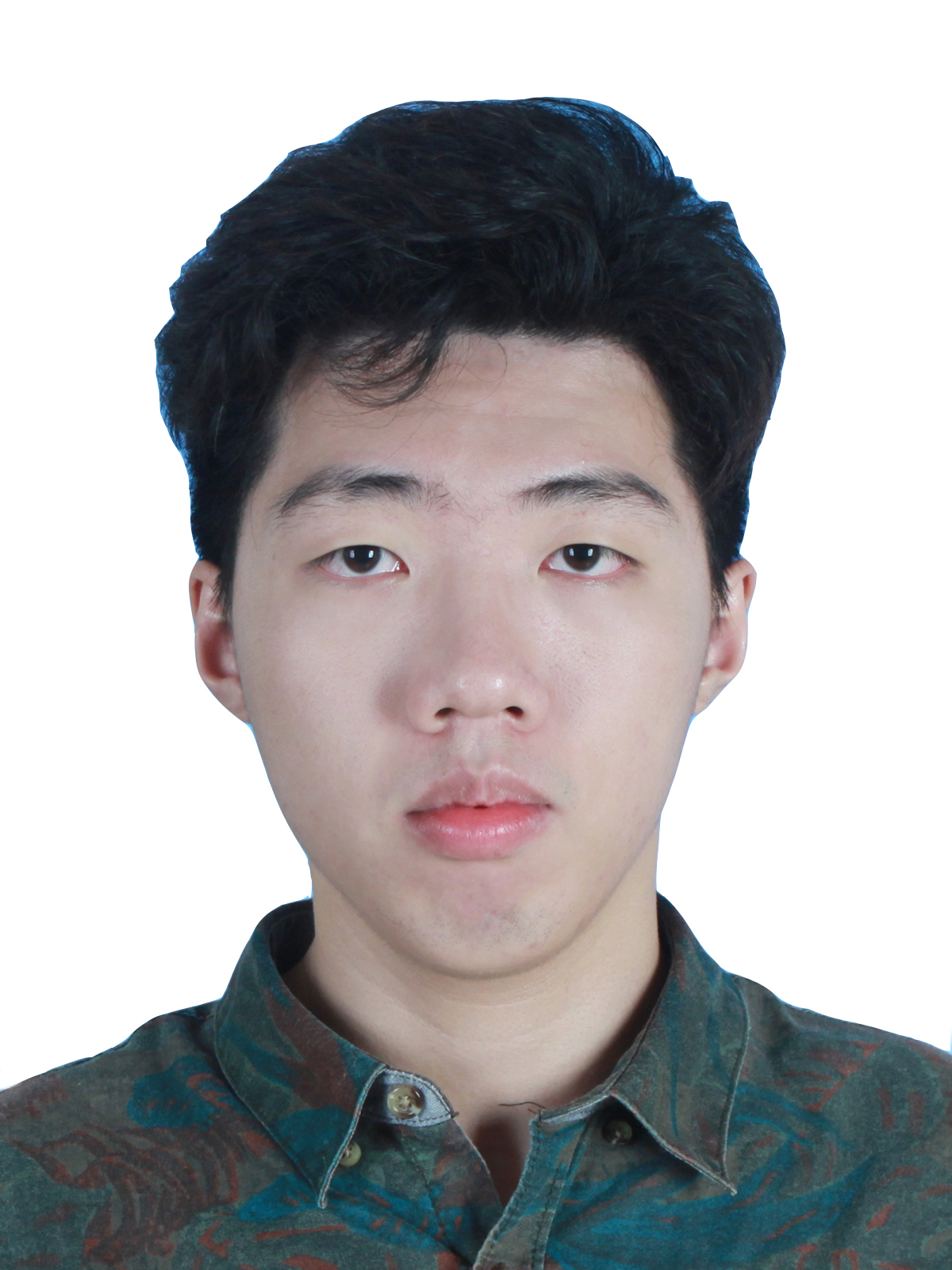}}]{Haofeng Li} received his B.S. degree from School of Data and Computer Science, Sun Yat-Sen University in 2015. He is currently pursuing the Ph.D. degree in the department of computer science, the University of Hong Kong. He is a recipient of Hong Kong PhD Fellowship. His current research interests include computer vision, image processing and deep learning.
\end{IEEEbiography}

\vspace{-10mm}
\begin{IEEEbiography}[{\includegraphics[width=1in,height=1.25in,clip,keepaspectratio]{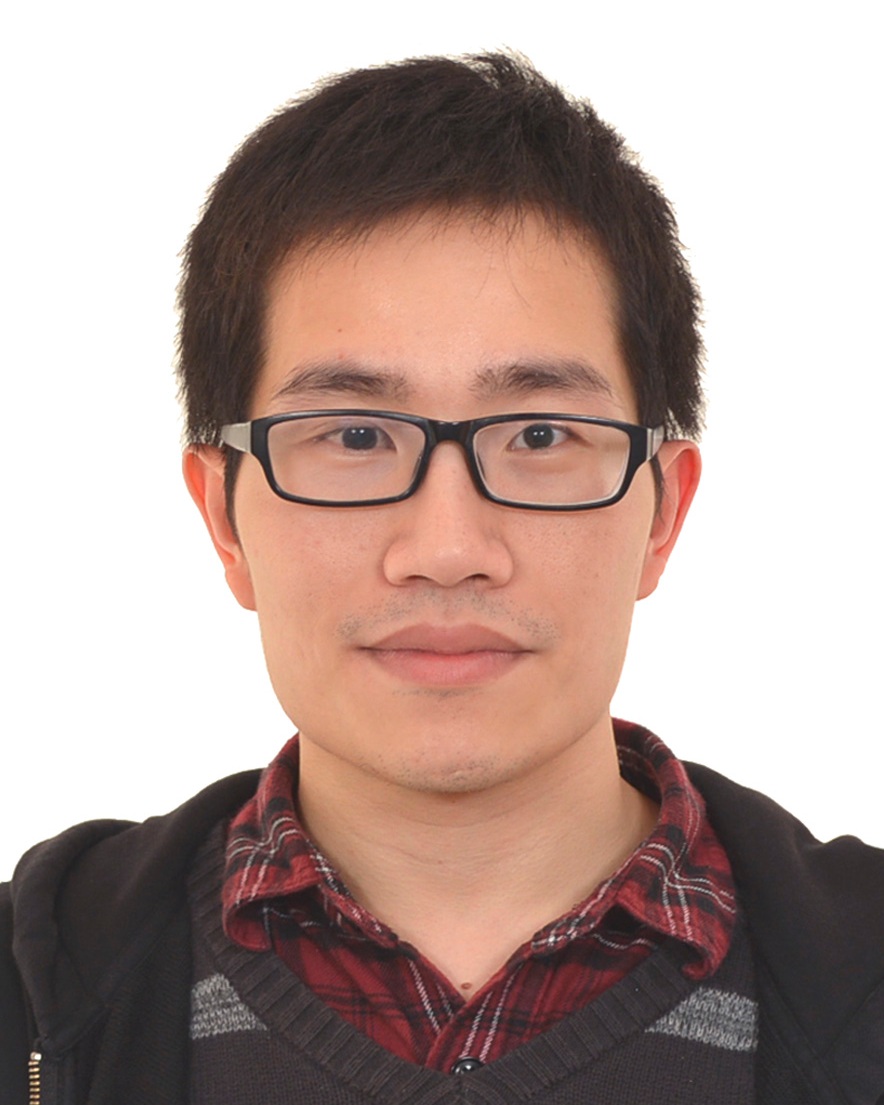}}]{Guanbin Li} (M'15) is currently an associate professor in School of Data and Computer Science, Sun Yat-sen University. He received his PhD degree from the University of Hong Kong in 2016. He was a recipient of Hong Kong PhD Fellowship. His current research interests include computer vision, image processing, and deep learning. He has authorized and co-authorized on more than 20 papers in top-tier academic journals and conferences. He serves as an area chair for the conference of VISAPP. He has been serving as a reviewer for numerous academic journals and conferences such as TPAMI, TIP, TMM, TC, CVPR, AAAI and IJCAI.
\end{IEEEbiography}

\vspace{-10mm}
\begin{IEEEbiography}[{\includegraphics[width=1in,height=1.25in,clip,keepaspectratio]{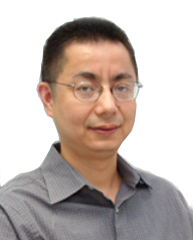}}]{Yizhou Yu} (M'10, SM'12, F'19) received the PhD degree from University of California at Berkeley in 2000. He is a professor at The University of Hong Kong, and was a faculty member at University of Illinois at Urbana-Champaign for twelve years. He is a recipient of 2002 US National Science Foundation CAREER Award, 2007 NNSF China Overseas Distinguished Young Investigator Award, and ACCV 2018 Best Application Paper Award. Prof Yu has served on the editorial board of IET Computer Vision, The Visual Computer, and IEEE Transactions on Visualization and Computer Graphics. He has also served on the program committee of many leading international conferences, including SIGGRAPH, SIGGRAPH Asia, and International Conference on Computer Vision. His current research interests include computer vision, deep learning, biomedical data analysis, computational visual media and geometric computing.
\end{IEEEbiography}

\end{document}